\documentclass{article}

\usepackage{microtype}
\usepackage{graphicx}
\usepackage{subfigure}
\usepackage{booktabs} 

\usepackage[accepted]{icml2025}
\usepackage{hyperref}

\usepackage{amsmath}
\usepackage{amssymb}
\usepackage{mathtools}
\usepackage{amsthm}

\usepackage[capitalize,noabbrev]{cleveref}

\theoremstyle{plain}
\newtheorem{theorem}{Theorem}[section]
\newtheorem{proposition}[theorem]{Proposition}

\newtheorem{corollary}[theorem]{Corollary}
\theoremstyle{definition}
\newtheorem{definition}[theorem]{Definition}

\theoremstyle{remark}

\usepackage[utf8]{inputenc} 
\usepackage[T1]{fontenc}  
\usepackage{hyperref}    
\usepackage{url}      
\usepackage{booktabs}    
\usepackage{amsfonts}    
\usepackage{nicefrac}    
\usepackage{microtype}   
\usepackage{xcolor}     

\DeclareMathOperator*{\E}{\mathbb{E}}
\usepackage{bm}
\usepackage{caption}
\usepackage{subcaption}
\usepackage[utf8]{inputenc}

\icmltitlerunning{Demystifying Spectral Bias on Real-World Data}
\author{%
  Itay Lavie \qquad Zohar Ringel \\
  Racah Institute of Physics\\
  Hebrew University of Jerusalem\\
  Jerusalem 91904, Israel  \\
  \texttt{\{itay.lavie, zohar.ringel\}@mail.huji.ac.il} \\
}
\title{}
\date{}

\begin{document}

\twocolumn[
\icmltitle{Demystifying Spectral Bias on Real-World Data}
\vspace{-1.5cm}
\maketitle
\vspace{-1.cm}






\icmlkeywords{Kernel Ridge Regression,Gaussian Process,NTK,NNGP,Deep Learning Theory,Symmetry,Sample Complexity,Learnability}

\vskip 0.3in
]




%

%

\begin{abstract}
Kernel ridge regression (KRR) and Gaussian processes (GPs) are fundamental tools in statistics and machine learning, with recent applications to highly over-parameterized deep neural networks. The ability of these tools to learn a target function is directly related to the eigenvalues of their kernel sampled on the input data distribution. Targets that have support on higher eigenvalues are more learnable. However, solving such eigenvalue problems on real-world data remains a challenge. Here, we consider cross-dataset learnability and show that one may use eigenvalues and eigenfunctions associated with highly idealized data measures to reveal spectral bias on complex datasets and bound learnability on real-world data. This allows us to leverage various symmetries that realistic kernels manifest to unravel their spectral bias.
\end{abstract}
\vspace{-0.8cm}
\section{Introduction}

Gaussian process regression (GPR) and kernel ridge regression (KRR) are workhorses of statistics and machine learning. GPR and KRR are also intimately related - given the same kernel function, they both result in the same predictor \citep{Rasmussen,kimeldorf_correspondence_1970}. 
More recently, several correspondences between GPs and kernel methods with deep neural networks (DNNs) have appeared \citep{lee2018deep,jacot2018neural,matthews2018gaussian,Novak2018BayesianDC}. Thus, apart from the direct interest in GPs and kernel methods, predicting what GPs and kernels can learn appears as a stepping stone to predicting neural networks' learnability. 

Learning in such models can be characterized by their so called spectral bias. A line of works~\citep{Silverman1984,Sollich2004,cohen_learning_2021,canatar_spectral_2021,bordelon2021spectrum,simon_eigenlearning_2023} assumes full knowledge of the to-be-learned target function and the underlying distribution of the data and studies the performance of the model, termed \emph{omniscient risk}~\cite{breiman_how_1983,wei_more_2022}. 
These works identify the eigenvalues and eigenfunctions associated with the kernel matrix in the infinite data limit as the main objects controlling learnability. In this eigen-learning framework, the regression target is decomposed to eigenfunctions and the regression roughly filters out eigenfunctions with eigenvalues below $\sigma^2/P$, $P$ being the amount of data and $\sigma^2$ the ridge parameter or an effective ridge parameter. Notwithstanding, this approach still leaves us with the formidable task of diagonalizing the kernel on the data measure. In addition to being a hard computational task, it also requires an accurate understanding of the underlying data measure, which is often out of our grasp. Thus, while kernels show much promise as indicators of DNN performance, leveraging them to provide insights into how DNNs perform on real-world data remains an open problem.

Under dataset \emph{symmetry} assumptions, the eigenlearning framework has been used to establish the spectral bias of different architectures. The spectrum of kernels associated with fully connected networks (FCNs) was found for rotationally symmetric data distributions, revealing a dimensionality curse, where the sample complexity scales as the input dimension to the polynomial power of the target~\citep{basri_convergence_2019,yang_fine-grained_2020,scetbon_spectral_2021,bietti_deep_2021}). Recently this strategy has been extended to transformers acting on permutation symmetric data~\citep{lavie_towards_2024}. Common to these works is the use of symmetry, namely, they rely on the known equivariance of the model which leads to a kernel symmetry, and assume this symmetry applies to the data as well. The result is a symmetry for the eigenvalue problem; representation theory arguments can then be directly applied to diagonalize/block-diagonalize kernels~\citep{fulton_representation_2004,tung_group_1985}. Furthermore, the finite sum rule on eigenvalues and their non-negativity in conjugation with degeneracies implied by symmetry further forces strict upper bounds on the eigenvalues (e.g. \citet{cohen_learning_2021}). Unfortunately, real-world data is rarely uniform or symmetric. Consequently, the eigenvalue problem, involving both the kernel and the data measure, loses its symmetry properties and the symmetry of the kernel (model equivariance) alone does not help one obtain the kernel spectrum required for characterizing the spectral bias via learnability.

In this work, we reveal the role played by the kernel (model) symmetry when acting on generic (real-world) data. Central to our bound is the use of an auxiliary test distribution ($q$), invariant under symmetries of the kernel, to measure learnability. We show that such \emph{cross-dataset learnability} can be bounded from above without ever solving the difficult kernel eigenvalue problem on the real data. Instead, our bound depends on two relatively accessible quantities: ({\bf 1}) The eigenfunctions and eigenvalues of the kernel on $q$ and ({\bf 2}) The norm of those eigenfunctions associated with the target on the real dataset. As we demonstrate, since the eigenvalue problem w.r.t. $q$ enjoys all kernel symmetries, it is largely tractable using representation theory tools. 
We further show that cross-dataset learnability can be used to bound the performance on the real dataset both from above and from below under assumptions on their importance weighting (see for example ~\citet{sugiyama_density_2012}). Finally, we find empirically that sample complexity predictions derived from our cross-dataset learnability correlate well with results on real datasets. 

\textbf{Our main contributions are}: 
\vspace{-0.4cm}
\begin{itemize}
\itemsep0em
  \item We prove an upper bound on the cross-dataset learnability of a feature, requiring minimal knowledge of the real data distribution - the norm of the target function and feature on the real data distribution.
  \item We give simple upper and lower bounds on the generalization performance on the real data given the cross-dataset learnability and the expected importance ratio.

  \item We derive sample complexity bound based on the cross-dataset learnability and show they correlate well with results on real-world datasets.

  \item We analyze spectral bias in transformers and show that our method predicts large sample complexity for copying heads, and hence for several elementary in-context learning mechanisms.
\end{itemize}

\subsection{Related Works}
\emph{Symmetry in kernels and neural networks \& spectral bias.} Symmetry has been used extensively to understand kernels and neural networks in the kernel regime from the perspective of spectral bias. Fully connected networks with data uniformly distributed on a hypersphere have been studied in \citet{basri_convergence_2019,bietti_deep_2021,scetbon_spectral_2021}. Data distributed non-uniformly on the hypersphere was studied in \citet{basri_frequency_2020}, while Gaussian data and data uniformly distributed on the hypercube were studied in \citet{yang_fine-grained_2020}. This line of study was extended beyond the natural rotation symmetry; \citet{bietti_sample_2021} studied how extending the symmetry group reduces the sample complexity, a result that was generalized further by \citet{tahmasebi_exact_2023}.
The eigenspectra of kernels corresponding to convolutional neural networks were studied in \citet{bietti_approximation_2021,xiao_eigenspace_2022,cagnetta_what_2023,geifman_spectral_2022}. Finally, \citet{lavie_towards_2024} studied transformers in the kernel regime by leveraging their permutation symmetry. Here, we show that spectral bias results have a universal component and show how they can be adapted from idealized symmetric datasets to real-world datasets. 

\emph{Learning \& multiple data distributions.} There is a large body of work on distributional shift and out-of-distribution generalization~\cite{ben-david_analysis_2006,ben-david_theory_2010,pan_survey_2010,sugiyama_machine_2012,zhao_learning_2019,arjovsky_out_2021,canatar_out--distribution_2021,ma_optimally_2023,feng_towards_2023}, however, this setting and its motivation are different from ours. In the study of distributional shifts, one assumes the training (source) distribution does not accurately reflect the test (target) distribution and tries to estimate (bound, guarantee) the performance on the test distribution, which is the true object of interest. This concept fundamentally differs from the setting in this work. Here we do not assume a difference between the underlying distribution for the test and train sets, rather, we are interested in the spectral bias on real-world datasets and use $q$, an auxiliary distribution, solely as a tool. Presenting $q$ allows us to capitalize on all the results mentioned in the previous paragraph about symmetry, even when the dataset does not respect those symmetries. We are aware of one previous work \cite{Opper1998} on GPR that predicts learning curves on arbitrary data, however, their setting is rather different from ours and assumes the target function itself is drawn from a Gaussian prior, and does not use symmetries or address the spectral bias of different models/kernels.
\vspace{-0.3cm}
\section{Cross-Dataset Learnability}
\label{sec:setting}
In this section, we present a short introduction to kernel ridge regression (KRR) or Gaussian process regression (GPR) and the concept of learnability, followed by a generalization to cross-dataset scenarios.

The regression setting includes a kernel function $k(x,y)$, a ridge parameter\footnote{or an effective ridge parameter is the case for neural tangent kernel~\citep{canatar_spectral_2021}} $\sigma^2$, and a dataset $D = \left\{(x_\mu,y(x_\mu)) \right\}_{\mu=1}^P$ of $P$ data points drawn i.i.d. from an underlying distribution $p$. $x_\mu$ denotes the $\mu$'th input and $y(x_\mu)$ is the regression target/label for $x_\mu$. In this case, the predictor is given by \vspace{-0.3cm}
\begin{equation}
  \hat{f}_{D}\left(x\right)=\sum_{\nu,\rho=1}^P k\left(x,x_{\nu}\right)\left[K+I\sigma^{2}\right]_{\nu\rho}^{-1}y(x_{\rho});\quad\left[K\right]_{\mu\nu}=k\left(x_{\mu},x_{\nu}\right),
\label{eq:predictor_on_data}
\end{equation}
where $I$ is the identity matrix and $x_\mu \in D$\footnote{The resulting predictor from KRR with a kernel function $k(x,y)$ and ridge $\delta$ is identical to GPR with covariance function $k(x,y)$ and observation uncertainty $\sigma^2 = \delta$.}. In a contemporary context, a kernel of particular interest is the neural tangent kernel (NTK) which describes an NN trained with gradient flow~\citep{jacot2018neural}. A second example is the neural network Gaussian process (NNGP) which describes Bayesian inference with a prior induced by the distribution of the NN weights at initialization~\citep{neal_priors_1996,lee2018deep} or when training an NN with noisy gradients~\citep{naveh_predicting_2021,welling_bayesian_2011}. 

We can define the \emph{learnability} of the target as 
\begin{equation}
  \mathcal{L} := \frac{\E_{x \sim p} \left[ y(x)\hat{f}_{D}(x) \right]}{\E_{x \sim p} \left[ y(x) y(x) \right]} 
\label{eq:population_learnability}
\end{equation}
The learnability monotonically increases with $P$ (the dataset size), and takes values in the range $[0,1)$ such that $\mathcal{L}\stackrel{P\to \infty}{\longrightarrow} 1$ for $y \in \mathcal{H}_k$ where $\mathcal{H}_k$ is the RKHS of $k$~\citep{simon_eigenlearning_2023}. Losses such as MSE can be expressed in terms of the learnability, in particular, when $\E_{x \sim p} \left[ y(x) y(x) \right]=1$ the population risk is simply $\operatorname{MSE}=(1-\mathcal{L})^2$.

The learnability is known to be controlled by the eigenvalue decomposition $\{(\psi_i(x),\eta_i) \}_{i=1}^\infty$ of the kernel operator in the infinite data limit $\hat{K}_p$
\begin{equation}
\begin{aligned}
  \hat{K}_{p} \psi_i (x) := \E_{x' \sim p} \left[ k(x,x') \psi_i(x') \right]= \eta_i \psi_i (x),
  \\
  \E_{x \sim p} \left[\psi_i(x)\psi_j(x) \right]=\delta_{ij},~~x\in \operatorname{supp}p,
\end{aligned}
\label{eq:eigenproblem}
\end{equation}
with $\delta_{ij}$ the Kronecker delta and $\E_{x \sim p} [ \,\cdot\,]$ being expectation value w.r.t the distribution $p$.

Under the equivalent kernel (EK) approximation~\cite{silverman_spline_1984,sollich_using_2004,cohen_learning_2021}, where essentially the regression is done of the full population rather than a specific realization of a dataset $D$, the learnability of a specific eigenfunction feature $\psi_i$ (setting $y(x)=\phi_i(x)$) takes an especially simple form
\begin{equation}
    \mathcal{L}_i = \frac{\E_{x \sim p} \left[ \psi_i(x)\hat{f}_{D}(x) \right]}{\E_{x \sim p} \left[ \psi_i(x) y(x) \right]} \stackrel{{\rm EK}}{\approx} \frac{\eta_i}{\eta_i+\sigma^2/P}.
\label{eq:eigenlearnability}
\end{equation}
The EK approximation is exact for large datasets $P \gg 1$ and large ridge $\sigma^2$. Beyond this regime, in high dimension, the same form of learnability holds when one replaces the "bare" ridge with an effective ridge $\sigma^2 \to \sigma^2_{\rm eff}$~\citep{simon_eigenlearning_2023,canatar_spectral_2021}.


Learnability results can be restated as a function of the number of samples, resulting in sample complexity; e.g. requiring learnability is $1-\epsilon$ \vspace{-0.3cm}
 \begin{equation}
   \mathcal{L}_i \stackrel{!}{=} 1-\epsilon \Rightarrow P^*_i = \eta_i^{-1} \sigma^2 \frac{1-\epsilon}{\epsilon},
 \end{equation}
with $P^*_i$ the sample complexity to achieve learnability that is $\epsilon$ close to unity for the feature (eigenfunction) $\psi_i$. Larger eigenvalues give better (lower) sample complexity, inducing a \emph{spectral bias} - the regression will learn those features first and use them to explain the data. Whether we need few or many samples, generalize or overfit, is largely dependent on the spectral bias and the support of the target on the eigenfunctions.

Even under the omniscient assumption, that is, assuming that the true population distribution and target function are known, solving the eigenvalue problem in Eq.\eqref{eq:eigenproblem} remains an intractable task, leaving the spectral bias on rich and complex data obscure. To bypass this difficulty, we consider here a cross-dataset generalization of this setting, where one solves the eigenvalue problem given in Eq.\eqref{eq:eigenproblem} on an auxiliary dataset with a probability measure $q$, but performs the regression on the dataset $D$. Analogous to the learnability in Eq.~\ref{eq:population_learnability}, we introduce \emph{cross-dataset learnability} where one replace the population distribution $p$ with the auxiliary distribution $q$ \vspace{-0.5cm}
\begin{equation}
  \mathcal{L}_i^{D,q} := \frac{\E_{x \sim q} \left[ \phi_{i}(x)\hat{f}_{D}(x) \right]}{\E_{x \sim q} \left[ \phi_{i}(x) y(x) \right]} 
\label{eq:our_learnability}
\end{equation} 
with $\phi_i$ an eigenfunction of the kernel w.r.t. $q$, namely, $\phi_i$ solves Eq.~\eqref{eq:eigenproblem} with $p$ replaced by $q$. The cross-dataset learnability reduces to the common learnability by simply choosing $q=p$.
We note that while the common learnability in Eqs.~(\ref{eq:population_learnability},\ref{eq:eigenlearnability}) is bounded $\mathcal{L}_i \in [0,1)$, the cross-dataset learnability is unbounded. As a consequence, maximizing cross-dataset learnability does not imply good learning; instead, one must require it to be close to unity, see also Eq.~\eqref{eq:importance_ratios}. In the next section, we present a tractable lower bound on the cross-dataset learnability and show empirically that it predicts spectral bias on real-world datasets. The rest of this section gives a more intuitive understanding of cross-dataset learnability and a formal connection to the population risk in terms of upper and lower bounds, based on expected density ratios.

\subsection{Interpreting Cross-Dataset Learnability}

Intuitively, cross-dataset learnability $\mathcal{L}_i^{D,q}$ is simply the ratio between the magnitude of the component $\phi_i$ in the predictor and what it should be to reconstruct the target function perfectly. The change of distribution amounts to learning from $D$ but judging how good the reconstruction is based on functional similarity on $q$, that is using $q$ as a test distribution. Choosing a simple $q$ has the advantage of providing a clear "ruler" for measuring network outputs, as demonstrated in the next example, perhaps capturing a notion of out-of-distribution generalization but the disadvantage of being uninformed about the details of the specific dataset. 

As an example, consider learning the parity function on the hypercube in dimension $d$ such that the target function $y(x)=\prod_{i=1}^d x_i$ for $x \in \{-1,1\}^d$. In the generic case, and even more so in under a uniform measure this is a notoriously hard learning task (e.g. ~\citet{shalev-shwartz_failures_2017,yang_fine-grained_2020}). However, if the data distribution $p$ correlated $x_i$'s perfectly and $d$ is even the target reduces to a constant $y(x)\equiv1$, making learning it trivial. In such case using a symmetric measure that is uniform over the hypercube, or its extension to the hypersphere makes it clear the function that performed perfectly on $p$ is not in fact parity, and cross-dataset learnability with uniform $q$ makes for better distribution to judge function similarity on.  We show that achieving good cross-dataset learnability for parity and any distribution on the hypercube with an auxiliary distribution that is uniform on the hypersphere requires sample complexity exponential in the dimension in Appendix~\ref{appendix:parity}.

\subsection{Cross-Dataset Learnability and Test Performance as Covariate Shift}
Covariate shift is a distributional shift scenario where the input distribution changes, but the relation between input and output stays the same~\citep{quinonero-candela_dataset_2009}. In such a scenario one commonly has a source dataset of inputs and outputs that can be used for training, but one is interested in the performance of a target (test) data, which comes from a different distribution. Cross-dataset learnability resembles a measure for performance under covariate shift, where we keep $q$ (the would-be target data) a degree of freedom. Here, the covariate shift is merely a tool to reveal the spectral bias of the kernel, without the need to solve the eigenvalue problem on $p$; where we are still interested in the performance on $p$ nonetheless. 

However, we may enjoy the rich literature on covariate shift (e.g. \citet{ben-david_analysis_2006,ben-david_theory_2010,zhao_learning_2019,mansour_domain_2023,ma_optimally_2023,feng_towards_2023}) to place bounds on the population risk ($p$) using the performance on $q$\footnote{That would most resemble an "inverse" covariate shift where one wants to bound performance on the source distribution by performance on the target distribution. Fortunately, many of the results in the literature are symmetric with respect to the source and target distributions.}. 
%
To make the paper self-contained we include here a simple result that bounds the population risk in terms of MSE from above and below by the cross-dataset learnability and the expected density ratios (also known as importance ratios)

\begin{proposition} 
\label{prop:covariate_shift}
Given the expected importance ratios defined in Eq.~\eqref{eq:importance_ratios}
\begin{equation}
    \bar{J}^{-1} \sum_i \left(1-\mathcal{L}^{D,q}_i\right)^2 \E_{x \sim q} [y(x) \phi_i (x)]^2
    \leq {\rm MSE}
\end{equation}
\begin{equation}
{\rm MSE}
    \leq
    \bar{I} \sum_i \left(1-\mathcal{L}^{D,q}_i\right)^2 \E_{x \sim q} [y(x) \phi_i (x)]^2.
\end{equation}
with ${\rm MSE}= \E_{x \sim p} \left[ \left(f(x)-y(x) \right)^2 \right]$
and $\bar{I},\,\bar{J}$ the expected density ratios \vspace{-0.2cm}
\begin{equation}
\begin{aligned}
    \bar{I} &:= \E_{x\sim p} I(x),~~\bar{J} &:= \E_{x\sim q} I^{-1}(x) ~~ &I (x) &:= \frac{p(x)}{q(x)}.
\end{aligned}
\label{eq:importance_ratios}
\end{equation}
The proof is given in Appendix~\ref{appendix:prop_proof}.
\end{proposition}

In practice, the importance ratio can be estimated from the empirical dataset, see~\cite{sugiyama_density_2012,kimura_short_2024} and references within.

\section{An (Almost) Training Set Universal Bound on Cross-Dataset Learnability}
\label{sec:results}
\begin{theorem}
\label{thm:x-ds_learnability}
 Given a GP kernel $k(x,y)$ and its eigendecomposition $\left\{ (\lambda_i,\phi_i(x))\right\}$ on an auxiliary probability density (measure) $q(x)$ and a dataset $D = \left\{(x_\mu,y(x_\mu)) \right\}_{\mu=1}^P$ of $P$ samples, 
 the cross-dataset learnability (see Eq.\ref{eq:our_learnability}) is bounded from above by
 \begin{equation}
   \mathcal{L}^{D,q}_i \leq \underbrace{\frac{\lambda_{i}P}{\sigma^{2}}}_\text{Universal}
   \underbrace{
   \frac{\sqrt{\E_{x\sim D}\left[\phi_{i}^{2}\left(x\right)\right]\E_{x\sim D}\left[y^{2} (x) \right]}}{\left| \E_{x \sim q(x)} [\phi_i(x) y(x)] \right|}}_\text{ Dataset and target dependent}.
\label{eq:learnability_bound_main}
 \end{equation}
with $\E_{x \sim D} [ \,\cdot\,]$\footnote{This quantity may naturally fluctuate a little as one scans the values of $P$.} ($\E_{x \sim q(x)}[ \,\cdot\,]$) being expectation value w.r.t the dataset $D$ ($q$). The bound holds as long as the training dataset $D$ is a subset of the support of $q$
\begin{equation}
  {\rm supp} (q) \supseteq \{x\}_{\mu=1}^P;
\label{eq:support_condition}
\end{equation} note that this is the only requirement on $q$ and one may choose the most favorable one within this class. The proof is given in Appendix~\ref{appendix:proof}.
\end{theorem}

We see that the information about the training dataset in~\eqref{eq:learnability_bound_main} is fully contained in the $\ell^2$ norm of the feature $\phi_i$ and the target $y$ on the dataset $D$. Save these norms, the spectral bias, as expressed by the learnability bound, is seen to be universal across datasets. 

The bound in Eq.~\ref{eq:learnability_bound_main} becomes vacuous for large $P$, as it can easily be larger than $1$, as required for perfect learnability. Nevertheless, as we show in Appendix~\ref{appendix:tylor} under favorable assumptions it is the optimal linear approximation and upper bound to the learnability at the small $P$ limit.

Our main use of the bound is predicting the sample complexity of different features and thus characterizing the spectral bias of the model

\begin{corollary}
As a corollary from~\ref{thm:x-ds_learnability} the number of samples $P^*$ required to achieve cross-dataset learnability $\mathcal{L}^{D,q}_t = 1-\epsilon$ for a specific target feature $\phi_t(x)$ is bounded from below by
\begin{equation}
  P^*
  \geq
  \sigma^{2} \lambda_{t}^{-1} (1-\epsilon) \frac{\left| \E_{x \sim q(x)} [\phi_t(x) y(x)] \right|}{\sqrt{\E_{x\sim D}\left[\phi_{t}^{2}\left(x\right)\right]\E_{x\sim D}\left[y^{2} (x) \right]}},
\label{eq:main_result}
\end{equation} 
\end{corollary}

The result in Eq.\eqref{eq:main_result} can be interpreted as follows. At least $P^*$ samples are required to learn a feature $\phi_i(x)$ from the dataset $D$, where the eigenvalue/features are found by performing eigenvalue decomposition of the kernel on $q$. This bound can be seen as a prescription for carrying over sample complexity / spectral bias results found on a favorable auxiliary measure $q$ to a rich dataset of interest $D$. It can be used whenever one is able to perform the eigenvalue decomposition on $q$ more easily than on $D$, for $q$ that satisfies Eq.\eqref{eq:support_condition}. 
%

Importantly, one is free to choose a favorable measure $q$. In particular one may choose $q$ such that it respects all the symmetries of the kernel function $k$, we dedicate the next subsection to investigate that.

\subsection{Symmetries and the auxiliary measure $q$}
Here, we motivate choosing the auxiliary distribution $q$ based on guiding principles of symmetry. 
\begin{definition}[Kernel Symmetry]
We say a kernel $k(x,y)$ has a symmetry group $G$ if
\begin{equation}
  \forall g \in G \quad k(T_g x, T_g y) = k(x,y)
\end{equation}
for $T_g$ a faithful representation of $g \in G$.
\emph{Intuitively}, a kernel symmetry means the kernel treats a pair of inputs and their symmetry augmented pairs in the same way. 
\end{definition}
\begin{definition}[Dataset Symmetry]
We say a dataset measure $q$ is symmetric under the action of a group $G$ if
\begin{equation}
  \forall g \in G \quad q(T_g x) = q(x)
\end{equation}
for $T_g$ a faithful representation of $g \in G$. \emph{Intuitively}, a dataset symmetry means a symmetry-augmented version is as likely to be seen in the dataset as the original one. 
\end{definition}

\begin{figure}[t!]
  \centering
    \includegraphics[width=0.49\textwidth]{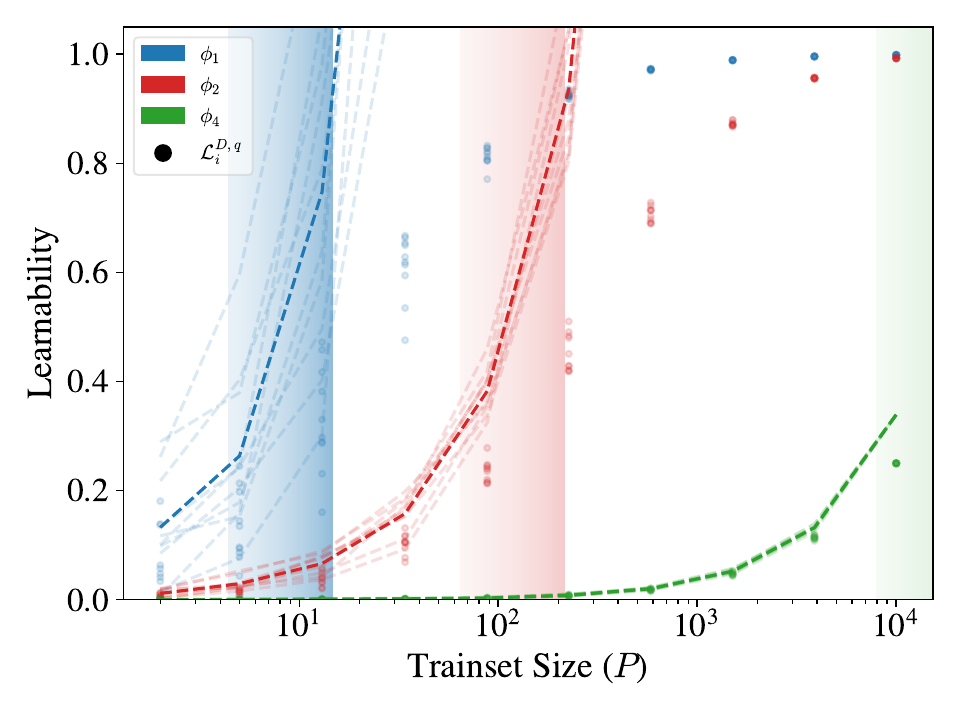}
    \vspace{-0.8cm}
    \caption{{\bf (The onset of learnability is tightly bounded in an idealized setting)} The cross-dataset learnability (dots) and our bound on the cross-dataset learnability (dashed) of a random linear $\phi_1$, quadratic $\phi_2$ and cubic $\phi_4$ target features. The trainset consists of $10^4$ samples drawn uniformly on the hypersphere $\mathbb{S}^{7}$ and $q$ is a uniform (continuous) distribution on the hypersphere. The shaded areas indicate a learning region, given by our bound taken at equality for $\epsilon\in[0,0.7]$. The bound is seen to be tight before and around the onset of learning even for a single realization. Notably, we do not expect the bound to be tight when the feature is already learned well, but to predict the minimum required number of samples for learning.}
  \label{fig:HS} \vspace{-0.3cm}
\end{figure}
When both the kernel and dataset respect a symmetry group $G$ we say it is a symmetry of the kernel \emph{operator}. A symmetry of the kernel operator can be used to asymptotically bound the eigenvalues from above using the dimension of their corresponding irreducible representations (irreps), as shown for the case of fully connected networks 
\citep{basri_convergence_2019,yang_fine-grained_2020,scetbon_spectral_2021,bietti_deep_2021}
and transformers \citep{lavie_towards_2024}. A natural choice for $q$ is therefore one that respects all the kernel symmetries.
\begin{equation}
  \forall g \in G \quad k(T_g x, T_g y) = k(x,y) \rightarrow q(T_g x) = q(x).
\end{equation}

In the last section, we analyze the spectral bias of transformers from the lens of symmetry and show large context size and large vocabulary imply strong spectral bias, predicting his sample complexity for the copying head task.

\begin{figure*}[t]
    \centering
\includegraphics[width=.68\columnwidth]{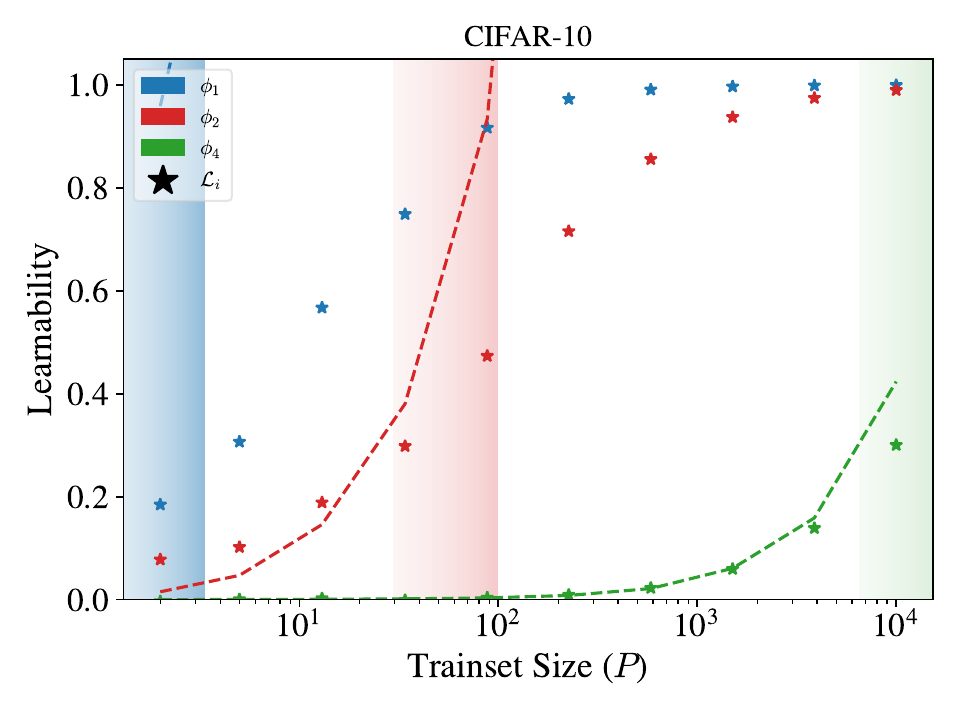}
\vspace{-0.35cm}\hfill
\includegraphics[width=.68\columnwidth]{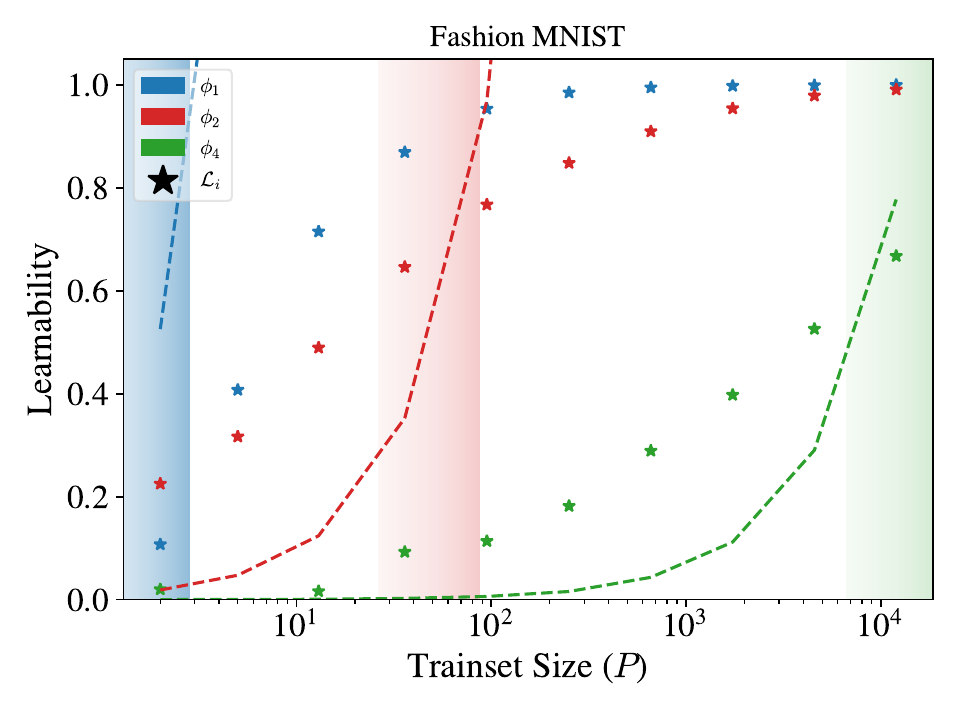}
\vspace{-0.35cm}\hfill
\includegraphics[width=.68\columnwidth]{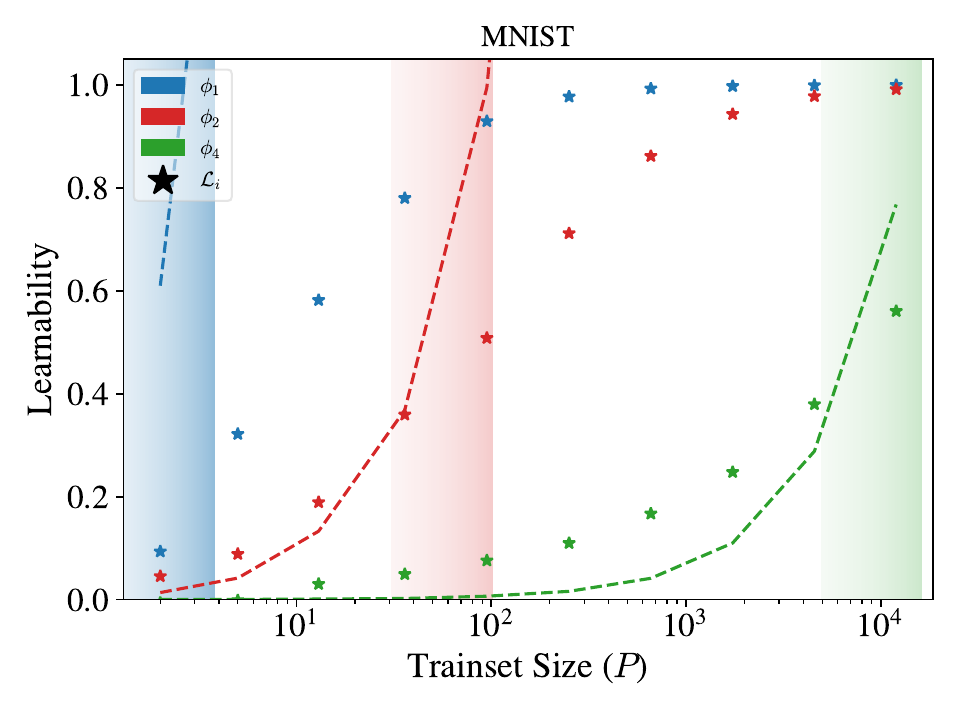}
\vspace{0.7cm} 
     \vspace{-0.5 cm}
  \caption{{\bf (Theory predicts spectral bias on real-world datasets)} The (test) learnability (dots) together with the bound on the cross-dataset learnability bound in Eq.~\eqref{eq:learnability_bound_main} (dashed). The shaded learning region indicated values of $P$ given by the bound in Eq.~\eqref{eq:main_result} for $0\leq\epsilon\leq0.7$. In most cases, the dashed bound and shaded learning regions give a good estimation of the sample complexity of the features.}
  \label{fig:spectral_bias_datasets} \vspace{-0.3cm}
\end{figure*}

\section{Experiments}
Here we experiment with cross-dataset learnability and spectral bias. We start by showing our bound on cross-dataset learnability in Eq.~\eqref{eq:learnability_bound_main} is tight during the onset of learning. We then move the real-world datasets: CIFAR-10~\citep{krizhevsky_learning_2009}, Fashion MNIST~\citep{DBLP:journals/corr/abs-1708-07747} and MNIST~\citep{lecun2010mnist} and show our sample complexity bound in Eq.~\eqref{eq:main_result} is able to predict spectral bias on them. Finally, we show that cross-dataset learnability tracks the learnability even more closely when we use PCA whitening to make the data and the auxiliary distribution more similar.

In Figure \ref{fig:HS} we show an experiment of exact KRR with a kernel that corresponds to a single hidden-layer ReLU network learning random linear, quadratic, and quartic target features $y=\phi_i$ for randomly chosen hyperspherical harmonics $\phi_1,\phi_2,\phi_4$ respectively. The dataset $D$ is $10^4$ samples drawn uniform i.i.d. on the hypersphere $\mathbb{S}^{7}$ ($d=8$). The symmetric auxiliary measure $q$ is naturally chosen to be the underlying symmetric distribution, (continuous) uniform on the hypersphere. We plot the cross-dataset learnability as in Eq.~\eqref{eq:our_learnability} (dots) together with our bound in Eq.~\eqref{eq:learnability_bound_main} (dashed line for each realization, with their average highlighted). In this case, the bound on the learnability is seen to approximate the beginning of the learning stage well. The regime in which the bound is tight indicates an important feature of our result. It captures the onset of learning and thus can be used to judge sample complexity; the bound misses the saturation of the learnability at later stages of learning. We stress that the dots indicate a single random realization of the dataset, and the bound is guaranteed to hold for every such realization.

Figure \ref{fig:spectral_bias_datasets}, shows the learnability\footnote{Calculated with the testset as a proxy for the population.} of random linear, quadratic, and quartic features $\phi_1,\phi_2,\phi_4$ for CIFAR-10, Fashion MNIST, and MNIST (stars).
The input dimension is reduced by PCA to $d=18$, approximately matching their intrinsic dimension~\citep{pope_intrinsic_2021,aumuller_role_2021} and capturing $\approx 80\%$ of the variance in the data; the data is then centered and scaled such that it is supported on the unit hypersphere.
We indicate the sample complexity prediction by taking the bound in Eq.~\eqref{eq:main_result} as equality for $\epsilon \in [0,0.7]$ and shading this area with matching colors. We also include the bound on cross-dataset learnability as a dashed line, as can be seen, our bound and the learnability intersect within the shaded learning region in most cases. Spectral bias is seen to be universal across datasets: $\phi_1$ is learned before $\phi_2,\phi_4$ and there is a predictable sample complexity gap between each one of the features. Additionally, we may conclude that it is unlikely that the classifier learned by an FCN-GP for any of these datasets include a high-degree polynomial, unless its norm on the dataset is exceedingly larger compared to its norm on the full hypersphere.

\begin{figure*}[t]
    \centering
\includegraphics[width=.68\columnwidth]{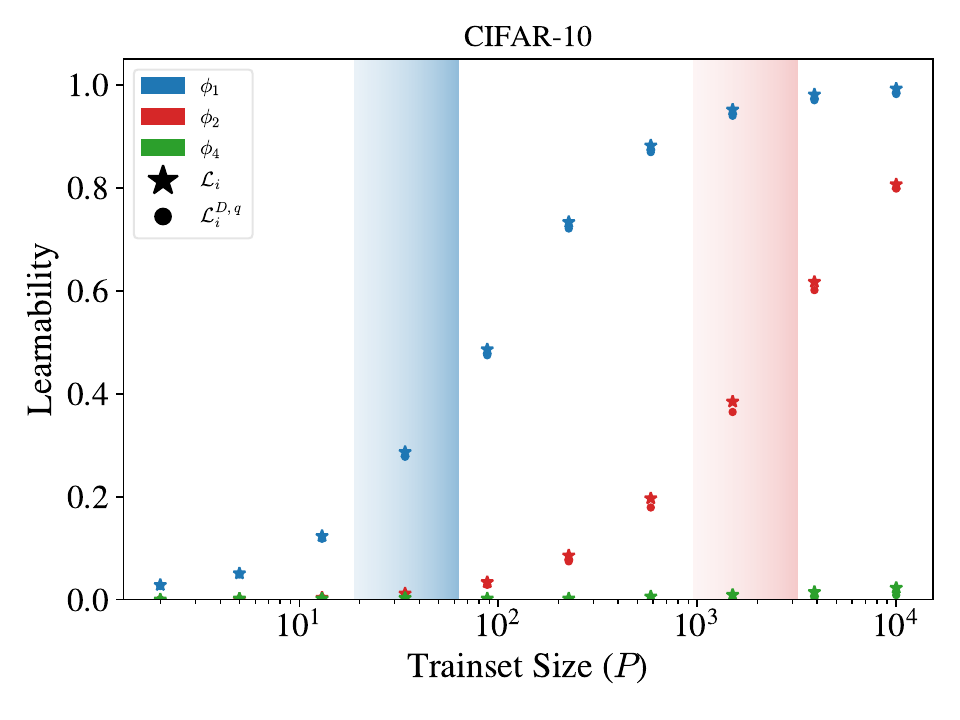}
\vspace{-0.35cm}\hfill
\includegraphics[width=.68\columnwidth]{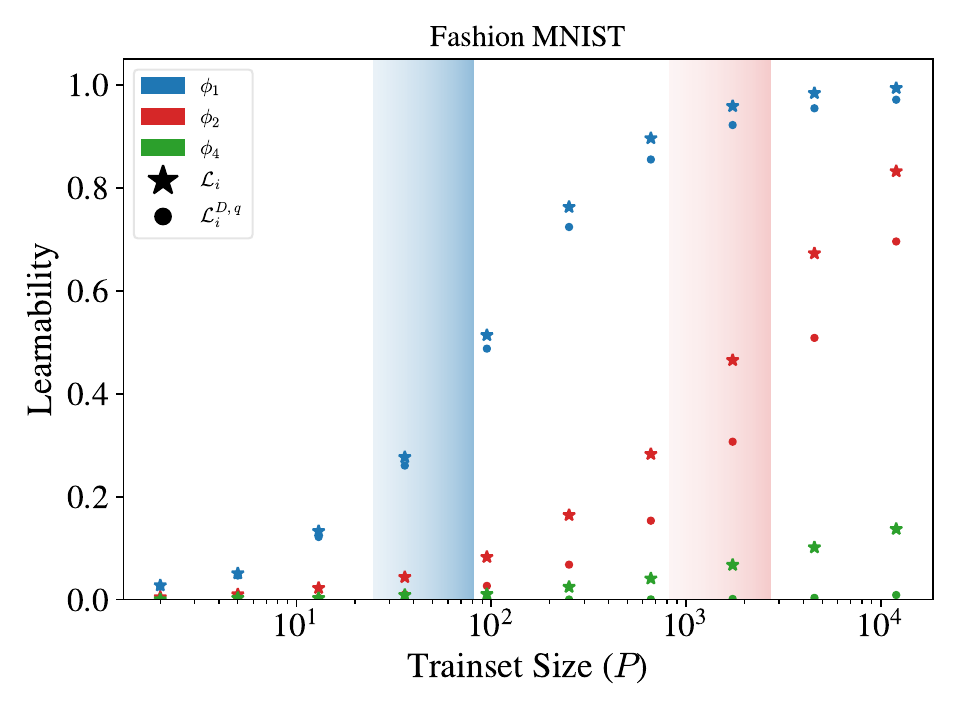}
\vspace{-0.35cm}\hfill
\includegraphics[width=.68\columnwidth]{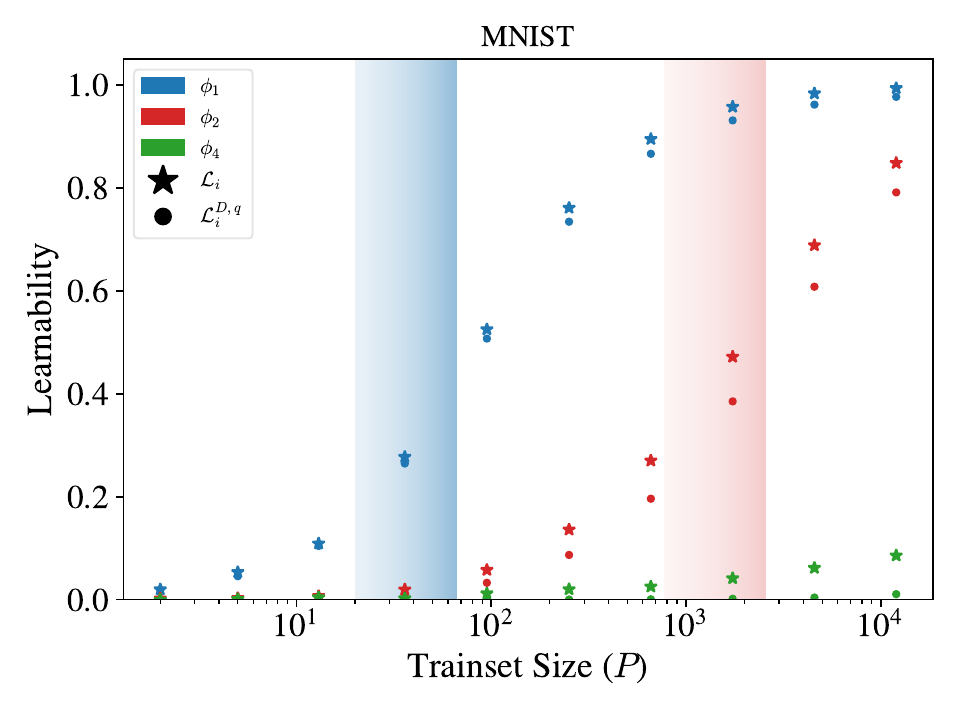}
\vspace{0.7cm} 
     \vspace{-0.5 cm}
    \caption{{\bf (Cross-dataset learnability approximates the learnability)} When the auxiliary distribution $q$ is similar to the data distribution the cross-dataset learnability (dots) approximates the learnability (stars).
    We use PCA whitening to bring the datasets' (CIFAR-10, Fashion MNIST, MNIST) distributions closer to the auxiliary distribution $q$ (uniform on the hypersphere $\mathbb{S}^17$).
    The shaded learning regions give a good indication of the sample complexity of the features.}
  \label{fig:spectral_bias_white_datasets} \vspace{-0.3cm}
\end{figure*}

Finally, we examine the choice of an idealized symmetric measure $q$ and motivate choices of $q$ that minimally change the data yet make the eigenvalue problem tractable (e.g. symmetries of the kernel). To this end, we show that when choosing similar distributions one enjoys a further benefit: the cross-dataset learnability indeed approximates the learnability well. We repeat the setting above but use PCA \emph{whitening}. PCA whitening makes the data covariance spherically symmetric hence making it more similar to the uniform distribution on the hypersphere. The results in Fig.~\ref{fig:spectral_bias_white_datasets} indeed show the cross-data set learnability (dots) approximated the learnability (stars) well. It is also worth commenting that dimensional reduction with PCA together with whitening was found to be highly beneficial to the performance of the NNGP and NTK of neural networks in~\citet{lee_finite_2020}.
\section{Vignettes}
\label{sec:vignettes}
Here we give two examples of implementing our main results in Eqs.~(\ref{eq:learnability_bound_main},\ref{eq:main_result}). The first one is a simple linear regression, showing how restricting the inputs the the sub-manifold that is relevant to the target can reduce sample complexity and how this fact enters our bound. The second example is inspired by in-context learning (ICL) in transformers and uses spectral bias to estimate the sample complexity of the copying head task.
\subsection{Awareness to Dataset \& Linear Regression on a Low Dimensional Data Manifold}
\label{subsec:vignette1}
The purpose of this example is twofold, first, present a simple setting as an example of how one can use our main result, and second, discuss the ways in which our bound accounts for different training datasets.

Consider the kernel $K(x,y)=\frac{1}{d} x\cdot y$ with $x,y \in R^d$ and $q(x)=\prod_{i=1}^d p_N(x_i)$ where $p_N(x_i)$ are standard centered Gaussians. Let $D$ consist of $P$ $d$-dimensional vectors sampled i.i.d. from the distribution $p$ such that $\vec{x} \sim p$ obeys
\begin{equation}
\vec{x} = (x_1,x_2,...,x_d),~~x_1 \sim N(0,d),x_{i>1}=0 .
\end{equation} Where the scale was chosen such that the sum of eigenvalues (trace) of the kernel on both measures equals unity
\begin{equation}
\intop k(\vec{x},\vec{x}) 
\frac{e^{-\frac{x_1^2}{2d}}}{\sqrt{2 \pi d}}
\prod_{i=2}^d \delta(x_i)\, d\vec{x}
=
\intop k(\vec{x},\vec{x}) 
\prod_{i=1}^d \frac{e^{-\frac{x_i^2}{2}}}{\sqrt{2 \pi}}\, d\vec{x} = 1
\end{equation}
with $\delta (x)$ the Dirac delta distribution, such that the learnability budget is the same in both distributions. Finally, let $y(\vec{x}) = x_1$. 

Let us first estimate the learnability here without using the above bound. The kernel on this dataset coincides with $d^{-1} x'_1 \cdot x_1$. From an EK perspective, this has a single non-zero eigenvalue $\eta$, associated with the function $\psi = x_1/\sqrt{d}$ (normalized w.r.t. $p$), given by 
\begin{align}
\intop_{-\infty}^{
\infty} 
dx'_1 \frac{x'_1 x_1}{d} \frac{x'_1}{\sqrt{d}}\frac{e^{-\frac{{x'}_1^2}{2d}}}{\sqrt{2 \pi d}}&= \eta \frac{x_1}{\sqrt{d}}; \quad \quad
\eta = 1 
\end{align}
hence the standard learnability (under EK approximation) is
\begin{equation}
    \mathcal{L} =\frac{\eta}{\eta + P/\sigma^2}. 
\end{equation}
Implying that $P^* = \eta^{-1} \sigma^2= \sigma^2$, for $\epsilon=1/2$, the crossover value that allows to learn half the target value. 

Next, we calculate the cross-dataset. To this end, we require the eigenfunctions and eigenvalues with respect to $q$. These can be checked to be  \vspace{-0.1cm}
\begin{align}
\phi_i(x) &= x_i \quad \quad \quad \quad \lambda_i = d^{-1}
\end{align}
Our target in these terms is $y(x) = \phi_1(x)$. Using EK approximation to estimate the predictor we find the cross-dataset learnability is \vspace{-0.1cm}
\begin{equation}
    \mathcal{L}^{p,q} =\frac{\eta}{\eta + P/\sigma^2}
\end{equation}
Notably, this result is exactly equal to the true learnability, hence implying the sample sample complexity $P^* = \sigma^2$.

Finally, we apply our bound.  Applying the bound in equation \eqref{eq:main_result} for $\epsilon=1/2$
\begin{align}
P^*
=
\frac{\sigma^{2} \lambda_{1}^{-1}}{2} \frac{\E_{x \sim q(x)} [\phi_1(x) y(x)]}{\sqrt{\E_{x\sim D}\left[\phi_{1}^{2}\left(x\right)\right]\E_{z\sim D}\left[y^{2} (x) \right]}}
\approx \frac{\sigma^2}{2} \frac{d}{d} = \frac{\sigma^2}{2}
\end{align}
where the last approximation is due to replacing empirical sampling ($D$) with the expected one. In this case, the resulting bound is seen to be within a factor of $2$ from the exact result.

\emph{Awareness to the training dataset.} Clearly, learning the target from the symmetric measure $q$ is a harder task, requiring $P^* = \sigma^2 d$ samples. We see that our bound \eqref{eq:main_result} encodes the information about the training dataset $D$ by the expected norms $\sqrt{\E_{x\sim D}\left[\phi_{1}^{2}\left(x\right)\right]},\sqrt{\E_{x\sim D}\left[y^{2} (x) \right]}$ of the target function and feature; which scales with $d$ in the above example.


\subsection{Learning Copying Heads with Transformers}
\label{subsec:vignette3}
Here, we use our theory to examine a concrete learning problem that has attracted recent attention~\citep{reddy_mechanistic_2023,edelman_evolution_2024,sanford_one-layer_2024,nichani_how_2024,singh_transient_2024} - learning induction heads~\citep{olsson_-context_2022}. An induction head performs an elementary form of ICL, where the next token is predicted based on the frequency of the tokens that followed it in the context, as in the form \vspace{-0.3cm}
\begin{equation}
    [A][B]...[A]\to[B].
\end{equation}
\citet{olsson_-context_2022} observed that learning induction heads consistently requires $\approx 10^9$ tokens (training steps) across a range of model sizes, with constant vocabulary size and context size. 

The large number of tokens required to learn this simple function might appear surprising. Motivated by this we give here a spectral bias perspective on the problem. We consider here the simpler problem of a Transformer kernel learning a copying head (that is simply copying the token that came before), which is essential for the creation of induction heads and mesa-optimization algorithms ~\citep{olsson_-context_2022,von_oswald_uncovering_2023}. Thus, such a simple example can be used as a lower bound on sample complexity for a wide variety of ICL mechanisms in the kernel regime. 

We define the input as $[X]_i^a$ where $i=1,...,V+1$ in the vocabulary index, and $a=1,...,L+1$ is the position of the token in the sequence. In this notation the target is $[Y(X)]^a_i=[X]^{a-1}_i$. For simplicity, we introduce a vector notation where $\vec{x}^a$ is a slice of $X$ at the position $a$, and a scalar notation $x^a_i = [X]^a_i$. We note that we do not consider causal masking in this example, as the target function, as defined in the simplified setup above, does not rely on a notion of causality. 

We set to establish a sample complexity lower bound for this task based on our bound in Eq.~\eqref{eq:main_result}. As an auxiliary distribution $q$ we choose a simple symmetric distribution where all tokens within sequences and samples are drawn uniformly i.i.d. from the vocabulary and are one-hot encoded, i.e. integers $v=1,...,V+1$ such that $x_i^a = \delta_{i,v}$ where $\delta$ is the Kronecker delta. The training dataset $D$ will be one-hot encoded, with the same vocabulary size $V+1$, but with arbitrary dependencies between the tokens. In particular, $D$ can be a true natural language dataset, tokenized with vocabulary size $V+1$.

The target function $Y(X)$ can be shown to include the feature (see Appendix \ref{appendix:copying_heads})
\begin{equation}
\begin{aligned}
  \vec{\phi}_t^a (X) = \frac{1}{z} \left( \vec{x}^{\,a-1} - \frac{1}{L}\sum_{b=1}^L \vec{x}^{\,b} - \frac{1}{V} \right) ; \\
  z = \sqrt{L \left( 1-L^{-1}+L^{-1}V^{-1} \right)},
\end{aligned}
\end{equation}
with an eigenvalue that can be bounded from above $\lambda_t
 \leq \frac{\E_{X \sim q}[k(X,X)]}{(L-1)(V-1)}$. We next use this result together with our main result~\eqref{eq:main_result} to bound the sample complexity of copying heads.

Under the assumption of one-hot encoded input $\E_{X\sim D}\left[{\rm Tr} [Y (X) Y^T (X)] \right] = \E_{X\sim D}\left[\sum_{a=1}^L \vec{y}^{\,a} (X) \cdot \vec{y}^{\,a} (X) \right]=L$ for all $D$. Finally 
\begin{equation}
\begin{aligned}
  &\E _{X\sim D}\left[{\rm Tr} [\Phi \left(X\right) \Phi^T \left(X\right) ] \right]
  =
  \E_{X\sim D}\left[\sum_{a=1}^L \vec{\phi}_{t}^{a}\left(X\right) \cdot \vec{\phi}_{t}^{a} \left(X\right) \right] \\ &=
  z^{-2} L \left(1-L^{-2}\sum_{a,b=1}^L \sum_{i=1}^V \E_{X \sim D} [x^a_i x^b_i] +V^{-1}\right)
\end{aligned}
\end{equation}
depends on the choice of $D$, nevertheless, one can easily derive bounds on the quantity. A simple bound is $\E _{X\sim D}\left[{\rm Tr} [\Phi \left(X\right) \Phi^T \left(X\right) ] \right] \leq z^{-2} L (1-L^{-1})+V^{-1}$ for any one-hot encoded input.

These results can be plugged in to~\eqref{eq:main_result} to find a general lower bound on the sample complexity
\begin{equation}
\begin{aligned}
  P^*
  &\geq 
  \sigma^{2} \lambda_t^{-1} (1-\epsilon) \frac{z}
  {\sqrt{\E _{X\sim D}\left[{\rm Tr} [\Phi \left(X\right) \Phi^T \left(X\right) ] \right]}}
  \\
  &\geq
  \sigma^{2} (1-\epsilon) \frac{(L-1)(V-1)}{\E_{X \sim q}[k(X,X)]}
  \frac{1-L^{-1}+L^{-1}V^{-1}}{\sqrt{1-L^{-1}+V^{-1}}}.
\end{aligned}
\end{equation}
Assuming a long context $L \gg 1$, large vocabulary $V \gg 1$, and normalized kernel $\E_{X \sim q}[k(X,X)] \simeq 1$, the sample complexity bound simply reads
\begin{equation}
  P^* \gtrsim \sigma^2 (1-\epsilon) L V.
\end{equation}
In simple terms, the number of samples has to scale like the product of the context length and the vocabulary size to learn copying heads. This result can be seen as a lower bound when models in the kernel limit can start performing ICL. We note \citet{olsson_-context_2022} used $V=2^{16}$ and $L=8192$ (giving $L\cdot V\approx0.5\cdot10^9$) showing that from a spectral bias perspective the sample complexity if such task is indeed very large. 

\section{Outlook}
\label{sec:outlook}
In this work, we have shown the spectral bias dictated by the model symmetries is largely universal, and that information about the dataset is accounted for by the change in the norm of the target function and features. Our results allow one to use the full symmetries of the kernel, even when they are not manifested in the training dataset and thus, naively, cannot be used. 

Kernel ridge regression and Gaussian process regression are not only well-motivated and studied frameworks but are also used in the study of neural networks through the NTK and NNGP correspondences~\citep{jacot2018neural,lee2018deep}. These correspondences have been used to characterize the spectral bias of different architectures such as fully connected networks ~\citep{basri_convergence_2019,yang_fine-grained_2020,scetbon_spectral_2021,bietti_deep_2021}, CNNs~\citep{bietti_approximation_2021,xiao_eigenspace_2022,cagnetta_what_2023} and transformers~\citep{lavie_towards_2024} under symmetry assumptions for the dataset. Our work generalizes these results to datasets that lack these symmetries. We further presented two examples using our bound: (1) an application of the bound for linear regression on a low dimensional manifold and (2) an application of the bound to the spectral bias of transformers, specifically giving a lower bound of the sample complexity of copying heads, a stepping stone for in-context learning. 

Much of the possibilities opened by our approach remain unexplored, like tightly bounding the sample complexity in cases where the target is very multi-spectral, which we elaborate on in Appendix~\ref{appendix:multi_spectral}. Another intriguing aspect is the application of such bound to ridgeless regression. When analyzing the dataset $D$, it has been shown there is an effective ridge coming from the unlearnable features~\citep{ cohen_learning_2021,canatar_spectral_2021,simon_eigenlearning_2023}. It would be of interest to find a similar effective ridge for our setting, using the eigenvalues and vectors from the symmetric distribution $q$, without having to solve the eigenvalue problem on the underlying data distribution $p$ (from which $D$ is drawn).
Finally, while our bound can be used to characterize what targets will not be learnable, a guarantee on the tightness of the bound can extend its applicability even further. 

\section*{Acknowledgments}
We thank Manfred Opper and Peter Sollich for the helpful discussions.
Z.R. and I.L. acknowledge support from ISF Grant 2250/19.
\newpage
\section*{Impact Statement}
This paper presents work whose goal is to advance the field of 
Machine Learning. There are many potential societal consequences 
of our work, none which we feel must be specifically highlighted here.
\bibliography{biblio,references}

\begin{thebibliography}{64}
\providecommand{\natexlab}[1]{#1}
\providecommand{\url}[1]{\texttt{#1}}
\expandafter\ifx\csname urlstyle\endcsname\relax
  \providecommand{\doi}[1]{doi: #1}\else
  \providecommand{\doi}{doi: \begingroup \urlstyle{rm}\Url}\fi

\bibitem[Arjovsky(2021)]{arjovsky_out_2021}
Arjovsky, M.
\newblock Out of {Distribution} {Generalization} in {Machine} {Learning}, March
  2021.
\newblock URL \url{http://arxiv.org/abs/2103.02667}.
\newblock arXiv:2103.02667.

\bibitem[Aumüller \& Ceccarello(2021)Aumüller and
  Ceccarello]{aumuller_role_2021}
Aumüller, M. and Ceccarello, M.
\newblock The role of local dimensionality measures in benchmarking nearest
  neighbor search.
\newblock \emph{Information Systems}, 101:\penalty0 101807, November 2021.
\newblock ISSN 0306-4379.
\newblock \doi{10.1016/j.is.2021.101807}.
\newblock URL
  \url{https://www.sciencedirect.com/science/article/pii/S0306437921000569}.

\bibitem[Basri et~al.(2019)Basri, Jacobs, Kasten, and
  Kritchman]{basri_convergence_2019}
Basri, R., Jacobs, D., Kasten, Y., and Kritchman, S.
\newblock The {Convergence} {Rate} of {Neural} {Networks} for {Learned}
  {Functions} of {Different} {Frequencies}.
\newblock In \emph{Advances in {Neural} {Information} {Processing} {Systems}},
  volume~32. Curran Associates, Inc., 2019.
\newblock URL
  \url{https://proceedings.neurips.cc/paper_files/paper/2019/hash/5ac8bb8a7d745102a978c5f8ccdb61b8-Abstract.html}.

\bibitem[Basri et~al.(2020)Basri, Galun, Geifman, Jacobs, Kasten, and
  Kritchman]{basri_frequency_2020}
Basri, R., Galun, M., Geifman, A., Jacobs, D., Kasten, Y., and Kritchman, S.
\newblock Frequency {Bias} in {Neural} {Networks} for {Input} of
  {Non}-{Uniform} {Density}.
\newblock In \emph{Proceedings of the 37th {International} {Conference} on
  {Machine} {Learning}}, pp.\  685--694. PMLR, November 2020.
\newblock URL \url{https://proceedings.mlr.press/v119/basri20a.html}.
\newblock ISSN: 2640-3498.

\bibitem[Ben-David et~al.(2006)Ben-David, Blitzer, Crammer, and
  Pereira]{ben-david_analysis_2006}
Ben-David, S., Blitzer, J., Crammer, K., and Pereira, F.
\newblock Analysis of {Representations} for {Domain} {Adaptation}.
\newblock In \emph{Advances in {Neural} {Information} {Processing} {Systems}},
  volume~19. MIT Press, 2006.
\newblock URL
  \url{https://papers.nips.cc/paper_files/paper/2006/hash/b1b0432ceafb0ce714426e9114852ac7-Abstract.html}.

\bibitem[Ben-David et~al.(2010)Ben-David, Blitzer, Crammer, Kulesza, Pereira,
  and Vaughan]{ben-david_theory_2010}
Ben-David, S., Blitzer, J., Crammer, K., Kulesza, A., Pereira, F., and Vaughan,
  J.~W.
\newblock A theory of learning from different domains.
\newblock \emph{Machine Learning}, 79\penalty0 (1):\penalty0 151--175, May
  2010.
\newblock ISSN 1573-0565.
\newblock \doi{10.1007/s10994-009-5152-4}.
\newblock URL \url{https://doi.org/10.1007/s10994-009-5152-4}.

\bibitem[Bietti(2021)]{bietti_approximation_2021}
Bietti, A.
\newblock Approximation and {Learning} with {Deep} {Convolutional} {Models}: a
  {Kernel} {Perspective}.
\newblock October 2021.
\newblock URL \url{https://openreview.net/forum?id=lrocYB-0ST2}.

\bibitem[Bietti \& Bach(2021)Bietti and Bach]{bietti_deep_2021}
Bietti, A. and Bach, F.
\newblock Deep {Equals} {Shallow} for {ReLU} {Networks} in {Kernel} {Regimes},
  August 2021.
\newblock URL \url{http://arxiv.org/abs/2009.14397}.
\newblock arXiv:2009.14397 [cs, stat].

\bibitem[Bietti et~al.(2021)Bietti, Venturi, and Bruna]{bietti_sample_2021}
Bietti, A., Venturi, L., and Bruna, J.
\newblock On the {Sample} {Complexity} of {Learning} under {Invariance} and
  {Geometric} {Stability}, November 2021.
\newblock URL \url{http://arxiv.org/abs/2106.07148}.
\newblock arXiv:2106.07148 [cs, stat].

\bibitem[{Blum} et~al.(2000){Blum}, {Kalai}, and {Wasserman}]{Blum2000}
{Blum}, A., {Kalai}, A., and {Wasserman}, H.
\newblock {Noise-Tolerant Learning, the Parity Problem, and the Statistical
  Query Model}.
\newblock \emph{arXiv e-prints}, art. cs/0010022, October 2000.
\newblock \doi{10.48550/arXiv.cs/0010022}.

\bibitem[Bordelon et~al.(2021)Bordelon, Canatar, and
  Pehlevan]{bordelon2021spectrum}
Bordelon, B., Canatar, A., and Pehlevan, C.
\newblock Spectrum dependent learning curves in kernel regression and wide
  neural networks.
\newblock 2021.

\bibitem[Breiman \& Freedman(1983)Breiman and Freedman]{breiman_how_1983}
Breiman, L. and Freedman, D.
\newblock How {Many} {Variables} {Should} be {Entered} in a {Regression}
  {Equation}?
\newblock \emph{Journal of the American Statistical Association}, 78\penalty0
  (381):\penalty0 131--136, 1983.
\newblock ISSN 0162-1459.
\newblock \doi{10.2307/2287119}.
\newblock URL \url{https://www.jstor.org/stable/2287119}.
\newblock Publisher: [American Statistical Association, Taylor \& Francis,
  Ltd.].

\bibitem[Cagnetta et~al.(2023)Cagnetta, Favero, and Wyart]{cagnetta_what_2023}
Cagnetta, F., Favero, A., and Wyart, M.
\newblock What {Can} {Be} {Learnt} {With} {Wide} {Convolutional} {Neural}
  {Networks}?
\newblock In \emph{Proceedings of the 40th {International} {Conference} on
  {Machine} {Learning}}, pp.\  3347--3379. PMLR, July 2023.
\newblock URL \url{https://proceedings.mlr.press/v202/cagnetta23a.html}.
\newblock ISSN: 2640-3498.

\bibitem[Canatar et~al.(2021{\natexlab{a}})Canatar, Bordelon, and
  Pehlevan]{canatar_out--distribution_2021}
Canatar, A., Bordelon, B., and Pehlevan, C.
\newblock Out-of-{Distribution} {Generalization} in {Kernel} {Regression}.
\newblock In \emph{Advances in {Neural} {Information} {Processing} {Systems}},
  November 2021{\natexlab{a}}.
\newblock URL \url{https://openreview.net/forum?id=-h6Ldc0MO-}.

\bibitem[Canatar et~al.(2021{\natexlab{b}})Canatar, Bordelon, and
  Pehlevan]{canatar_spectral_2021}
Canatar, A., Bordelon, B., and Pehlevan, C.
\newblock Spectral bias and task-model alignment explain generalization in
  kernel regression and infinitely wide neural networks.
\newblock \emph{Nature Communications}, 12\penalty0 (1):\penalty0 2914, May
  2021{\natexlab{b}}.
\newblock ISSN 2041-1723.
\newblock \doi{10.1038/s41467-021-23103-1}.
\newblock URL \url{https://www.nature.com/articles/s41467-021-23103-1}.
\newblock Number: 1 Publisher: Nature Publishing Group.

\bibitem[Cohen et~al.(2021)Cohen, Malka, and Ringel]{cohen_learning_2021}
Cohen, O., Malka, O., and Ringel, Z.
\newblock Learning curves for overparametrized deep neural networks: {A} field
  theory perspective.
\newblock \emph{Physical Review Research}, 3\penalty0 (2):\penalty0 023034,
  April 2021.
\newblock \doi{10.1103/PhysRevResearch.3.023034}.
\newblock URL \url{https://link.aps.org/doi/10.1103/PhysRevResearch.3.023034}.
\newblock Publisher: American Physical Society.

\bibitem[Edelman et~al.(2024)Edelman, Edelman, Goel, Malach, and
  Tsilivis]{edelman_evolution_2024}
Edelman, B.~L., Edelman, E., Goel, S., Malach, E., and Tsilivis, N.
\newblock The {Evolution} of {Statistical} {Induction} {Heads}: {In}-{Context}
  {Learning} {Markov} {Chains}, February 2024.
\newblock URL \url{http://arxiv.org/abs/2402.11004}.
\newblock arXiv:2402.11004 [cs].

\bibitem[Feng et~al.(2023)Feng, He, Wang, Wang, and Zhang]{feng_towards_2023}
Feng, X., He, X., Wang, C., Wang, C., and Zhang, J.
\newblock Towards a {Unified} {Analysis} of {Kernel}-based {Methods} {Under}
  {Covariate} {Shift}.
\newblock \emph{Advances in Neural Information Processing Systems},
  36:\penalty0 73839--73851, December 2023.
\newblock URL
  \url{https://proceedings.neurips.cc/paper_files/paper/2023/hash/e9b0ae84d6879b30c78cb8537466a4e0-Abstract-Conference.html}.

\bibitem[{Frye} \& {Efthimiou}(2012){Frye} and {Efthimiou}]{Frye2012}
{Frye}, C. and {Efthimiou}, C.~J.
\newblock {Spherical Harmonics in p Dimensions}.
\newblock \emph{arXiv e-prints}, art. arXiv:1205.3548, May 2012.
\newblock \doi{10.48550/arXiv.1205.3548}.

\bibitem[Fulton \& Harris(2004)Fulton and Harris]{fulton_representation_2004}
Fulton, W. and Harris, J.
\newblock \emph{Representation {Theory}: {A} {First} {Course}}, volume 129 of
  \emph{Graduate {Texts} in {Mathematics}}.
\newblock Springer, New York, NY, 2004.
\newblock ISBN 978-3-540-00539-1 978-1-4612-0979-9.
\newblock \doi{10.1007/978-1-4612-0979-9}.
\newblock URL \url{http://link.springer.com/10.1007/978-1-4612-0979-9}.

\bibitem[Geifman et~al.(2022)Geifman, Galun, Jacobs, and
  Ronen]{geifman_spectral_2022}
Geifman, A., Galun, M., Jacobs, D., and Ronen, B.
\newblock On the {Spectral} {Bias} of {Convolutional} {Neural} {Tangent} and
  {Gaussian} {Process} {Kernels}.
\newblock \emph{Advances in Neural Information Processing Systems},
  35:\penalty0 11253--11265, December 2022.
\newblock URL
  \url{https://proceedings.neurips.cc/paper_files/paper/2022/hash/48fd58527b29c5c0ef2cae43065636e6-Abstract-Conference.html}.

\bibitem[Jacot et~al.(2018)Jacot, Gabriel, and Hongler]{jacot2018neural}
Jacot, A., Gabriel, F., and Hongler, C.
\newblock Neural tangent kernel: Convergence and generalization in neural
  networks.
\newblock \emph{Advances in neural information processing systems}, 31, 2018.

\bibitem[Kimeldorf \& Wahba(1970)Kimeldorf and
  Wahba]{kimeldorf_correspondence_1970}
Kimeldorf, G.~S. and Wahba, G.
\newblock A {Correspondence} {Between} {Bayesian} {Estimation} on {Stochastic}
  {Processes} and {Smoothing} by {Splines}.
\newblock \emph{The Annals of Mathematical Statistics}, 41\penalty0
  (2):\penalty0 495--502, April 1970.
\newblock ISSN 0003-4851, 2168-8990.
\newblock \doi{10.1214/aoms/1177697089}.
\newblock URL
  \url{https://projecteuclid.org/journals/annals-of-mathematical-statistics/volume-41/issue-2/A-Correspondence-Between-Bayesian-Estimation-on-Stochastic-Processes-and-Smoothing/10.1214/aoms/1177697089.full}.
\newblock Publisher: Institute of Mathematical Statistics.

\bibitem[Kimura \& Hino(2024)Kimura and Hino]{kimura_short_2024}
Kimura, M. and Hino, H.
\newblock A {Short} {Survey} on {Importance} {Weighting} for {Machine}
  {Learning}, May 2024.
\newblock URL \url{http://arxiv.org/abs/2403.10175}.
\newblock arXiv:2403.10175 [cs].

\bibitem[Krizhevsky(2009)]{krizhevsky_learning_2009}
Krizhevsky, A.
\newblock Learning {Multiple} {Layers} of {Features} from {Tiny} {Images}.
\newblock 2009.
\newblock URL
  \url{https://www.semanticscholar.org/paper/Learning-Multiple-Layers-of-Features-from-Tiny-Krizhevsky/5d90f06bb70a0a3dced62413346235c02b1aa086}.

\bibitem[König(1986)]{konig_eigenvalue_1986}
König, H.
\newblock \emph{Eigenvalue {Distribution} of {Compact} {Operators}}, volume~16
  of \emph{Operator {Theory}: {Advances} and {Applications}}.
\newblock Birkhäuser, Basel, 1986.
\newblock ISBN 978-3-0348-6280-6 978-3-0348-6278-3.
\newblock \doi{10.1007/978-3-0348-6278-3}.
\newblock URL \url{http://link.springer.com/10.1007/978-3-0348-6278-3}.

\bibitem[Lavie et~al.(2024)Lavie, Gur-Ari, and Ringel]{lavie_towards_2024}
Lavie, I., Gur-Ari, G., and Ringel, Z.
\newblock Towards {Understanding} {Inductive} {Bias} in {Transformers}: {A}
  {View} {From} {Infinity}.
\newblock In \emph{Proceedings of the 41st {International} {Conference} on
  {Machine} {Learning}}, pp.\  26043--26069. PMLR, July 2024.
\newblock URL \url{https://proceedings.mlr.press/v235/lavie24a.html}.
\newblock ISSN: 2640-3498.

\bibitem[LeCun et~al.(2010)LeCun, Cortes, and Burges]{lecun2010mnist}
LeCun, Y., Cortes, C., and Burges, C.
\newblock {MNIST} handwritten digit database.
\newblock \emph{ATT Labs [Online]. Available:
  http://yann.lecun.com/exdb/mnist}, 2, 2010.

\bibitem[Lee et~al.(2018)Lee, Sohl-dickstein, Pennington, Novak, Schoenholz,
  and Bahri]{lee2018deep}
Lee, J., Sohl-dickstein, J., Pennington, J., Novak, R., Schoenholz, S., and
  Bahri, Y.
\newblock Deep neural networks as gaussian processes.
\newblock In \emph{International Conference on Learning Representations}, 2018.
\newblock URL \url{https://openreview.net/forum?id=B1EA-M-0Z}.

\bibitem[Lee et~al.(2020)Lee, Schoenholz, Pennington, Adlam, Xiao, Novak, and
  Sohl-Dickstein]{lee_finite_2020}
Lee, J., Schoenholz, S.~S., Pennington, J., Adlam, B., Xiao, L., Novak, R., and
  Sohl-Dickstein, J.
\newblock Finite {Versus} {Infinite} {Neural} {Networks}: an {Empirical}
  {Study}, September 2020.
\newblock URL \url{http://arxiv.org/abs/2007.15801}.
\newblock arXiv:2007.15801 [cs, stat].

\bibitem[Ma et~al.(2023)Ma, Pathak, and Wainwright]{ma_optimally_2023}
Ma, C., Pathak, R., and Wainwright, M.~J.
\newblock Optimally tackling covariate shift in {RKHS}-based nonparametric
  regression, June 2023.
\newblock URL \url{http://arxiv.org/abs/2205.02986}.
\newblock arXiv:2205.02986.

\bibitem[Mansour et~al.(2023)Mansour, Mohri, and
  Rostamizadeh]{mansour_domain_2023}
Mansour, Y., Mohri, M., and Rostamizadeh, A.
\newblock Domain {Adaptation}: {Learning} {Bounds} and {Algorithms}, November
  2023.
\newblock URL \url{http://arxiv.org/abs/0902.3430}.
\newblock arXiv:0902.3430 [cs].

\bibitem[Matthews et~al.(2018)Matthews, Rowland, Hron, Turner, and
  Ghahramani]{matthews2018gaussian}
Matthews, A. G. d.~G., Rowland, M., Hron, J., Turner, R.~E., and Ghahramani, Z.
\newblock Gaussian process behaviour in wide deep neural networks.
\newblock \emph{arXiv preprint arXiv:1804.11271}, 2018.

\bibitem[Naveh et~al.(2021)Naveh, Ben~David, Sompolinsky, and
  Ringel]{naveh_predicting_2021}
Naveh, G., Ben~David, O., Sompolinsky, H., and Ringel, Z.
\newblock Predicting the outputs of finite deep neural networks trained with
  noisy gradients.
\newblock \emph{Physical Review E}, 104\penalty0 (6):\penalty0 064301, December
  2021.
\newblock \doi{10.1103/PhysRevE.104.064301}.
\newblock URL \url{https://link.aps.org/doi/10.1103/PhysRevE.104.064301}.
\newblock Publisher: American Physical Society.

\bibitem[Neal(1996)]{neal_priors_1996}
Neal, R.~M.
\newblock Priors for {Infinite} {Networks}.
\newblock In Neal, R.~M. (ed.), \emph{Bayesian {Learning} for {Neural}
  {Networks}}, Lecture {Notes} in {Statistics}, pp.\  29--53. Springer, New
  York, NY, 1996.
\newblock ISBN 978-1-4612-0745-0.
\newblock \doi{10.1007/978-1-4612-0745-0_2}.
\newblock URL \url{https://doi.org/10.1007/978-1-4612-0745-0_2}.

\bibitem[Nichani et~al.(2024)Nichani, Damian, and Lee]{nichani_how_2024}
Nichani, E., Damian, A., and Lee, J.~D.
\newblock How {Transformers} {Learn} {Causal} {Structure} with {Gradient}
  {Descent}.
\newblock In \emph{Proceedings of the 41st {International} {Conference} on
  {Machine} {Learning}}, pp.\  38018--38070. PMLR, July 2024.
\newblock URL \url{https://proceedings.mlr.press/v235/nichani24a.html}.
\newblock ISSN: 2640-3498.

\bibitem[Novak et~al.(2018)Novak, Xiao, Bahri, Lee, Yang, Hron, Abolafia,
  Pennington, and Sohl-Dickstein]{Novak2018BayesianDC}
Novak, R., Xiao, L., Bahri, Y., Lee, J., Yang, G., Hron, J., Abolafia, D.~A.,
  Pennington, J., and Sohl-Dickstein, J.~N.
\newblock Bayesian deep convolutional networks with many channels are gaussian
  processes.
\newblock In \emph{International Conference on Learning Representations}, 2018.
\newblock URL \url{https://api.semanticscholar.org/CorpusID:57721101}.

\bibitem[Olsson et~al.(2022)Olsson, Elhage, Nanda, Joseph, DasSarma, Henighan,
  Mann, Askell, Bai, Chen, Conerly, Drain, Ganguli, Hatfield-Dodds, Hernandez,
  Johnston, Jones, Kernion, Lovitt, Ndousse, Amodei, Brown, Clark, Kaplan,
  McCandlish, and Olah]{olsson_-context_2022}
Olsson, C., Elhage, N., Nanda, N., Joseph, N., DasSarma, N., Henighan, T.,
  Mann, B., Askell, A., Bai, Y., Chen, A., Conerly, T., Drain, D., Ganguli, D.,
  Hatfield-Dodds, Z., Hernandez, D., Johnston, S., Jones, A., Kernion, J.,
  Lovitt, L., Ndousse, K., Amodei, D., Brown, T., Clark, J., Kaplan, J.,
  McCandlish, S., and Olah, C.
\newblock In-context {Learning} and {Induction} {Heads}, September 2022.
\newblock URL \url{http://arxiv.org/abs/2209.11895}.
\newblock arXiv:2209.11895 [cs].

\bibitem[Opper \& Vivarelli(1998)Opper and Vivarelli]{Opper1998}
Opper, M. and Vivarelli, F.
\newblock General bounds on bayes errors for regression with gaussian
  processes.
\newblock In Kearns, M., Solla, S., and Cohn, D. (eds.), \emph{Advances in
  Neural Information Processing Systems}, volume~11. MIT Press, 1998.
\newblock URL
  \url{https://proceedings.neurips.cc/paper_files/paper/1998/file/c7af0926b294e47e52e46cfebe173f20-Paper.pdf}.

\bibitem[Pan \& Yang(2010)Pan and Yang]{pan_survey_2010}
Pan, S.~J. and Yang, Q.
\newblock A {Survey} on {Transfer} {Learning}.
\newblock \emph{IEEE Transactions on Knowledge and Data Engineering},
  22\penalty0 (10):\penalty0 1345--1359, October 2010.
\newblock ISSN 1558-2191.
\newblock \doi{10.1109/TKDE.2009.191}.
\newblock URL
  \url{https://ieeexplore.ieee.org/document/5288526/?arnumber=5288526}.
\newblock Conference Name: IEEE Transactions on Knowledge and Data Engineering.

\bibitem[Pope et~al.(2021)Pope, Zhu, Abdelkader, Goldblum, and
  Goldstein]{pope_intrinsic_2021}
Pope, P., Zhu, C., Abdelkader, A., Goldblum, M., and Goldstein, T.
\newblock {THE} {INTRINSIC} {DIMENSION} {OF} {IMAGES} {AND} {ITS} {IMPACT} {ON}
  {LEARNING}.
\newblock 2021.

\bibitem[Quiñonero-Candela(2009)]{quinonero-candela_dataset_2009}
Quiñonero-Candela, J. (ed.).
\newblock \emph{Dataset shift in machine learning}.
\newblock Neural information processing series. MIT Press, Cambridge, Mass,
  2009.
\newblock ISBN 978-0-262-17005-5 978-0-262-54587-7.
\newblock OCLC: ocn227205909.

\bibitem[Rasmussen \& Williams(2006)Rasmussen and Williams]{Rasmussen}
Rasmussen, C.~E. and Williams, C. K.~I.
\newblock \emph{Gaussian processes for machine learning.}
\newblock Adaptive computation and machine learning. MIT Press, 2006.
\newblock ISBN 026218253X.

\bibitem[Reddy(2023)]{reddy_mechanistic_2023}
Reddy, G.
\newblock The mechanistic basis of data dependence and abrupt learning in an
  in-context classification task.
\newblock October 2023.
\newblock URL \url{https://openreview.net/forum?id=aN4Jf6Cx69}.

\bibitem[Sanford et~al.(2024)Sanford, Hsu, and
  Telgarsky]{sanford_one-layer_2024}
Sanford, C., Hsu, D., and Telgarsky, M.
\newblock One-layer transformers fail to solve the induction heads task, August
  2024.
\newblock URL \url{http://arxiv.org/abs/2408.14332}.
\newblock arXiv:2408.14332 [cs].

\bibitem[Scetbon \& Harchaoui(2021)Scetbon and
  Harchaoui]{scetbon_spectral_2021}
Scetbon, M. and Harchaoui, Z.
\newblock A {Spectral} {Analysis} of {Dot}-product {Kernels}.
\newblock In \emph{Proceedings of {The} 24th {International} {Conference} on
  {Artificial} {Intelligence} and {Statistics}}, pp.\  3394--3402. PMLR, March
  2021.
\newblock URL \url{https://proceedings.mlr.press/v130/scetbon21b.html}.
\newblock ISSN: 2640-3498.

\bibitem[Shalev-Shwartz et~al.(2017)Shalev-Shwartz, Shamir, and
  Shammah]{shalev-shwartz_failures_2017}
Shalev-Shwartz, S., Shamir, O., and Shammah, S.
\newblock Failures of {Gradient}-{Based} {Deep} {Learning}.
\newblock In \emph{Proceedings of the 34th {International} {Conference} on
  {Machine} {Learning}}, pp.\  3067--3075. PMLR, July 2017.
\newblock URL \url{https://proceedings.mlr.press/v70/shalev-shwartz17a.html}.
\newblock ISSN: 2640-3498.

\bibitem[Silverman(1984{\natexlab{a}})]{Silverman1984}
Silverman, B.~W.
\newblock {Spline Smoothing: The Equivalent Variable Kernel Method}.
\newblock \emph{The Annals of Statistics}, 12\penalty0 (3):\penalty0 898 --
  916, 1984{\natexlab{a}}.
\newblock \doi{10.1214/aos/1176346710}.
\newblock URL \url{https://doi.org/10.1214/aos/1176346710}.

\bibitem[Silverman(1984{\natexlab{b}})]{silverman_spline_1984}
Silverman, B.~W.
\newblock Spline {Smoothing}: {The} {Equivalent} {Variable} {Kernel} {Method}.
\newblock \emph{The Annals of Statistics}, 12\penalty0 (3):\penalty0 898--916,
  1984{\natexlab{b}}.
\newblock ISSN 0090-5364.
\newblock URL \url{https://www.jstor.org/stable/2240968}.
\newblock Publisher: Institute of Mathematical Statistics.

\bibitem[Simon et~al.(2023)Simon, Dickens, Karkada, and
  Deweese]{simon_eigenlearning_2023}
Simon, J.~B., Dickens, M., Karkada, D., and Deweese, M.
\newblock The {Eigenlearning} {Framework}: {A} {Conservation} {Law}
  {Perspective} on {Kernel} {Ridge} {Regression} and {Wide} {Neural}
  {Networks}.
\newblock \emph{Transactions on Machine Learning Research}, February 2023.
\newblock ISSN 2835-8856.
\newblock URL \url{https://openreview.net/forum?id=FDbQGCAViI}.

\bibitem[Singh et~al.(2024)Singh, Chan, Moskovitz, Grant, Saxe, and
  Hill]{singh_transient_2024}
Singh, A., Chan, S., Moskovitz, T., Grant, E., Saxe, A., and Hill, F.
\newblock The transient nature of emergent in-context learning in transformers.
\newblock \emph{Advances in Neural Information Processing Systems}, 36, 2024.
\newblock URL
  \url{https://proceedings.neurips.cc/paper_files/paper/2023/hash/58692a1701314e09cbd7a5f5f3871cc9-Abstract-Conference.html}.

\bibitem[Sollich \& Williams(2004{\natexlab{a}})Sollich and
  Williams]{Sollich2004}
Sollich, P. and Williams, C.
\newblock Using the equivalent kernel to understand gaussian process
  regression.
\newblock In Saul, L., Weiss, Y., and Bottou, L. (eds.), \emph{Advances in
  Neural Information Processing Systems}, volume~17. MIT Press,
  2004{\natexlab{a}}.
\newblock URL
  \url{https://proceedings.neurips.cc/paper_files/paper/2004/file/d89a66c7c80a29b1bdbab0f2a1a94af8-Paper.pdf}.

\bibitem[Sollich \& Williams(2004{\natexlab{b}})Sollich and
  Williams]{sollich_using_2004}
Sollich, P. and Williams, C.
\newblock Using the {Equivalent} {Kernel} to {Understand} {Gaussian} {Process}
  {Regression}.
\newblock In \emph{Advances in {Neural} {Information} {Processing} {Systems}},
  volume~17. MIT Press, 2004{\natexlab{b}}.
\newblock URL
  \url{https://proceedings.neurips.cc/paper/2004/hash/d89a66c7c80a29b1bdbab0f2a1a94af8-Abstract.html}.

\bibitem[Sugiyama \& Kawanabe(2012)Sugiyama and
  Kawanabe]{sugiyama_machine_2012}
Sugiyama, M. and Kawanabe, M.
\newblock \emph{Machine {Learning} in {Non}-{Stationary} {Environments}:
  {Introduction} to {Covariate} {Shift} {Adaptation}}.
\newblock The MIT Press, March 2012.
\newblock ISBN 978-0-262-30122-0.
\newblock \doi{10.7551/mitpress/9780262017091.001.0001}.
\newblock URL
  \url{https://direct.mit.edu/books/monograph/3774/Machine-Learning-in-Non-Stationary}.

\bibitem[Sugiyama et~al.(2012)Sugiyama, Suzuki, and
  Kanamori]{sugiyama_density_2012}
Sugiyama, M., Suzuki, T., and Kanamori, T.
\newblock \emph{Density ratio estimation in machine learning}.
\newblock Cambridge University Press, New York, 2012.
\newblock ISBN 978-1-139-23325-5 978-1-139-03561-3.

\bibitem[Tahmasebi \& Jegelka(2023)Tahmasebi and Jegelka]{tahmasebi_exact_2023}
Tahmasebi, B. and Jegelka, S.
\newblock The {Exact} {Sample} {Complexity} {Gain} from {Invariances} for
  {Kernel} {Regression}.
\newblock \emph{Advances in Neural Information Processing Systems},
  36:\penalty0 55616--55646, December 2023.

\bibitem[Tung(1985)]{tung_group_1985}
Tung, W.-K.
\newblock \emph{Group {Theory} in {Physics}: {An} {Introduction} to {Symmetry}
  {Principles}, {Group} {Representations}, and {Special} {Functions} in
  {Classical} and {Quantum} {Physics}}.
\newblock WORLD SCIENTIFIC, August 1985.
\newblock ISBN 978-9971-966-57-7 978-981-238-498-0.
\newblock \doi{10.1142/0097}.
\newblock URL \url{http://www.worldscientific.com/worldscibooks/10.1142/0097}.

\bibitem[von Oswald et~al.(2023)von Oswald, Niklasson, Schlegel, Kobayashi,
  Zucchet, Scherrer, Miller, Sandler, Arcas, Vladymyrov, Pascanu, and
  Sacramento]{von_oswald_uncovering_2023}
von Oswald, J., Niklasson, E., Schlegel, M., Kobayashi, S., Zucchet, N.,
  Scherrer, N., Miller, N., Sandler, M., Arcas, B. A.~y., Vladymyrov, M.,
  Pascanu, R., and Sacramento, J.
\newblock Uncovering mesa-optimization algorithms in {Transformers}, September
  2023.
\newblock URL \url{http://arxiv.org/abs/2309.05858}.
\newblock arXiv:2309.05858 [cs].

\bibitem[Wei et~al.(2022)Wei, Hu, and Steinhardt]{wei_more_2022}
Wei, A., Hu, W., and Steinhardt, J.
\newblock More {Than} a {Toy}: {Random} {Matrix} {Models} {Predict} {How}
  {Real}-{World} {Neural} {Representations} {Generalize}, March 2022.
\newblock URL \url{http://arxiv.org/abs/2203.06176}.
\newblock arXiv:2203.06176 [cs, stat].

\bibitem[Welling \& Teh(2011)Welling and Teh]{welling_bayesian_2011}
Welling, M. and Teh, Y.~W.
\newblock Bayesian learning via stochastic gradient langevin dynamics.
\newblock In \emph{Proceedings of the 28th {International} {Conference} on
  {International} {Conference} on {Machine} {Learning}}, {ICML}'11, pp.\
  681--688, Madison, WI, USA, June 2011. Omnipress.
\newblock ISBN 978-1-4503-0619-5.

\bibitem[Xiao et~al.(2017)Xiao, Rasul, and
  Vollgraf]{DBLP:journals/corr/abs-1708-07747}
Xiao, H., Rasul, K., and Vollgraf, R.
\newblock Fashion-{MNIST}: a novel image dataset for benchmarking machine
  learning algorithms.
\newblock \emph{CoRR}, abs/1708.07747, 2017.
\newblock URL \url{http://arxiv.org/abs/1708.07747}.
\newblock arXiv: 1708.07747 tex.bibsource: dblp computer science bibliography,
  https://dblp.org tex.timestamp: Mon, 13 Aug 2018 16:47:27 +0200.

\bibitem[Xiao(2022)]{xiao_eigenspace_2022}
Xiao, L.
\newblock Eigenspace {Restructuring}: {A} {Principle} of {Space} and
  {Frequency} in {Neural} {Networks}.
\newblock In \emph{Proceedings of {Thirty} {Fifth} {Conference} on {Learning}
  {Theory}}, pp.\  4888--4944. PMLR, June 2022.
\newblock URL \url{https://proceedings.mlr.press/v178/xiao22a.html}.
\newblock ISSN: 2640-3498.

\bibitem[Yang \& Salman(2020)Yang and Salman]{yang_fine-grained_2020}
Yang, G. and Salman, H.
\newblock A {Fine}-{Grained} {Spectral} {Perspective} on {Neural} {Networks},
  April 2020.
\newblock URL \url{http://arxiv.org/abs/1907.10599}.
\newblock arXiv:1907.10599 [cs, stat].

\bibitem[Zhao et~al.(2019)Zhao, Combes, Zhang, and Gordon]{zhao_learning_2019}
Zhao, H., Combes, R. T.~D., Zhang, K., and Gordon, G.
\newblock On {Learning} {Invariant} {Representations} for {Domain}
  {Adaptation}.
\newblock In \emph{Proceedings of the 36th {International} {Conference} on
  {Machine} {Learning}}, pp.\  7523--7532. PMLR, May 2019.
\newblock URL \url{https://proceedings.mlr.press/v97/zhao19a.html}.
\newblock ISSN: 2640-3498.

\end{thebibliography}
\bibliographystyle{icml2025}
\newpage

\newpage
\appendix

\section{Proof of The Main Theorem}
\label{appendix:proof}

The main idea of the proof builds upon the fact that while the eigendecomposition of the operator 
\begin{equation}
  \hat{K}_q \phi_i (x) = \intop k(x,x') \phi_i(x') q(x') dx' = \lambda_i \phi_i (x)
\end{equation}
depends on the measure $q$, the representation of the kernel found by Mercer decomposition holds for all datasets within the support on $q$. We then use a known Mercer decomposition on $q$ to represent the kernel function that is sampled in the predictor $\hat{f}_D$ from $D$.

The usual GPR/KRR predictor on the dataset $D$ is given by Eq.\eqref{eq:predictor_on_data}.
Using Mercer's theorem~\citep{konig_eigenvalue_1986} to decompose the kernel function to eigenfunctions \emph{on $q$}, the predictor can be written as
\begin{align}
  \hat{f}_{D}\left(x\right) &= \sum_{\nu\rho=1}^P\sum_{i=1}^{\infty}\phi_{i}\left(x\right)\lambda_{i}\phi_{i}\left(x_{\nu}\right)\left[K+I\sigma^{2}\right]_{\nu\rho}^{-1}y(x_{\rho});
  \nonumber \\
  \hat{K}_q \phi_{i}(x) &= \lambda_{i}\phi_{i}\left(x\right).
\end{align}
We now project the predictor onto the target feature $\phi_t$ with the inner product $\langle \cdot,\cdot \rangle_q$ defined with $q(x)$ as a weighting function (such that $\langle f,g \rangle_q = \E_{x \sim q} [f(x) g(x)]$)
\begin{align}
    \left\langle \phi_{t},\hat{f}_{D}\right\rangle _{q}&=\sum_{\nu\rho}\sum_{i=1}^{\infty}\left\langle \phi_{t},\phi_{i}\right\rangle _{q}\lambda_{i}\phi_{i}\left(x_{\nu}\right)\left[K+I\sigma^{2}\right]_{\nu\rho}^{-1}y(x_{\rho})
    \nonumber \\
    &=\lambda_{t}\sum_{\nu\rho}\phi_{t}\left(x_{\nu}\right)\left[K+I\sigma^{2}\right]_{\nu\rho}^{-1}y(x_{\rho}),
\end{align}
where the inner product can be carried out immediately based on the orthonormality of $\{ \phi_i \}_i$ w.r.t. the inner product $\langle \cdot,\cdot \rangle_q$. We may now use Cauchy-Schwartz inequality to bound the inner product on $D$, given by the summation on the index $\nu$,
\begin{equation}
\begin{aligned}
  \left| \left\langle \phi_{t},\hat{f}_{D}\right\rangle _{q} \right| \leq \frac{\lambda_{t}}{\sigma^2}&\sqrt{\sum_{\nu}\phi^2_{t}\left(x_{\nu}\right)} 
  \\
  &\cdot \sqrt{\sum_{\mu} \left( \sum_\rho \left[\sigma^{-2}K+I\right]_{\mu\rho}^{-1}y (x_{\rho}) \right)^2}.
\end{aligned}
\end{equation}
Lastly, since $\left[\left(\sigma^{-2}K+I\right)\right]^{-1}$ is weakly contracting we can bound the result from above by
\begin{align}
  \left| \left\langle \phi_{t},\hat{f}_{D}\right\rangle _{q} \right| \leq \sigma^{-2} \lambda_{t}\sqrt{\left(\sum_{\nu}\phi_{t}\left(x_{\nu}\right)\phi_{t}\left(x_{\nu}\right)\right)\left(\sum_{\mu}y_{\mu}y_{\mu}\right)} 
  \nonumber \\
  = \sigma^{-2} \lambda_{t}P\sqrt{\E_{x\sim D}\left[\phi_{t}^{2}\left(x\right)\right]\E_{x\sim D}\left[y^{2} (x) \right]}.
\end{align}
plugging this result into the definition of cross-dataset learnability (Eq.~\eqref{eq:our_learnability}) and noting we can choose the sign of $\langle \phi_t,y \rangle$ to be positive without loss of generality we arrive at the result Eq.~\eqref{eq:learnability_bound_main} in Thm.~\ref{thm:x-ds_learnability}.
From here the corollary is a straightforward progress:
To quantify whether the feature is learned or not we require that the learnability is $\epsilon$ close to perfect $\mathcal{L}^{D,q}_i \stackrel{!}{=}1-\epsilon$ yielding the result in Eq.\ref{eq:main_result}.

\section{Comparing Learnability Bound to Our Cross-Dataset Learnability Bound}
\label{appendix:tylor}
Our bound \eqref{eq:learnability_bound_main}
can be compared with a common learnability measure at the EK limit
that can be found only given a solution to the generically intractable eigenvalue problem on the data distribution $p$.
Expanding the learnability in Eq.~\eqref{eq:eigenlearnability} away from $1$ we find  
\begin{equation}
  \mathcal{L}_{i} = \frac{\eta_{i}}{\eta_{i}+\sigma^{2}/P} = \frac{\eta_{i} P}{\sigma^2} + O\left( \left(\frac{\eta_{i} P}{\sigma^2} \right)^2 \right) \leq \frac{\eta_{i} P}{\sigma^2},
  \label{eq:data_dep_bound}
\end{equation}
where the last inequality arises because the learnability is a concave function of $P$. Like Eq.~\eqref{eq:learnability_bound_main}, Eq.~\eqref{eq:data_dep_bound} is a necessary condition for learnability but it uses a different pair of eigenvalue-eigenfunction $\eta_t,\psi_t$. $\psi_t$ is guaranteed to be normalized as the eigendecomposition was carried out with the same probability measure as the dataset. From this perspective, the bound can be expected to be tight when $P \ll \eta_{i}^{-1} \sigma^2$ and less tight as $P$ grows larger. This is coherent with the view of our bound as an \emph{necessary} condition for learnability, rather than an exact prediction of the learnability. The power expansion approximates the learnability well when far from good learnability and bounds it from above throughout training. 
\section{Proof of Proposition \ref{prop:covariate_shift}}
\label{appendix:prop_proof}
We would like to show that
\begin{equation}
\begin{aligned}
    \bar{J}^{-1} \sum_i \left(1-\mathcal{L}^{D,q}_i\right)^2 \E_{x \sim q} [y(x) \phi_i (x)]^2
    \leq {\rm MSE}
\\
{\rm MSE}
    \leq
    \bar{I} \sum_i \left(1-\mathcal{L}^{D,q}_i\right)^2 \E_{x \sim q} [y(x) \phi_i (x)]^2.
\end{aligned}
\end{equation}
with
\begin{equation}
    \bar{I} := \E_{x\sim p} I(x),~~\bar{J} := \E_{x\sim q} I^{-1}(x), ~~ I (x) := \frac{p(x)}{q(x)}.
\end{equation}
we start with the first inequality
\begin{equation}
\begin{aligned}
    \int (f - y)^2 q(x) dx \leq \int |f - y| q(x) dx = \int |f - y| \frac{q(x)}{p(x)} p(x) dx
    \\
    \leq \int (f - y)^2 p(x) dx\int \left(\frac{q(x)}{p(x)}\right)^2 p(x) dx = \bar{J} \int (f - y)^2 p(x) dx
\end{aligned}
\end{equation}
where the first inequality arises from the fact the $\ell^1$ is larger than the $\ell^2$ norm (taken w.r.t. q), and the second inequality is Cauchy-Schwartz w.r.t. to p.
the second inequality is achieved in the same way but starting from the population loss on $p$.
Finally from the definition of cross-dataset learnability and the orthogonality of the eigenfunctions 
\begin{equation}
    \int (f - y)^2 q(x) dx=\sum_i \left(1-\mathcal{L}^{D,q}_i\right)^2 \E_{x \sim q} [y(x) \phi_i (x)]^2
\end{equation}
\section{Multi-spectral extensions}
\label{appendix:multi_spectral}
In the main text, we bound the magnitude of the predictor's projection onto the kernel's eigenfunctions on $q$ and compare it to those of the target. Here we consider the case where $y(x)$ is highly multispectral and receives contributions from a large, potentially infinite, number of $\phi_t$ modes. In such a scenario, $\left\langle \phi_{t},y \right\rangle _{q}^2$ would scale inversely with the number of dominant modes whereas $y^2$ appearing in our bound would remain $O(1)$. As a consequence, the bound may become very loose as is the case where the data distribution and $q$ match, where our bound would be $O \left(\left\langle \phi_{t},y\right\rangle _{q}^{-2} \right)$ off the EK result. 

Here we extend our bound to certain multispectral circumstances. Specifically, let us assume that $y(x)$ can be written as $y_<(x)+y_>(x)$ such that $y_>(x)$ is spanned by $\{\phi_i(x) \}_i$ having $\lambda_i \leq \lambda_>$. Furthermore let us assume $|y|^2 \propto |y_>(x)|^2 \propto O(1)$ and $|y_>(x)|_q^2 \geq O(1)$. As we argue next in such cases we essentially derive a similar bound with $\lambda_>$ playing the role of $\lambda_t$. Specifically we consider $|\langle \hat{f}_D (x),y_>(x)\rangle|_q$ given by 
\begin{align}
|\int dx &q(x) y_>(x) \sum_{\mu \nu}K(x,x_{\mu})[K+I \sigma^2]^{-1}_{\mu \nu} y_{\nu}|  
\nonumber  \\
&= |\sum_{t\geq t_>} \lambda_t \left\langle \phi_{t},y\right\rangle_{q} \phi_t(x_{\mu})[K+I \sigma^2]^{-1}_{\mu \nu} y_{\nu}| \nonumber  \\ 
&= 
|[\sum_{t \geq t_>} \lambda_t \left\langle \phi_{t},y\right\rangle_{q} \vec{\phi}_t]^T[K+I \sigma^2]^{-1} \vec{y}| 
 \nonumber \\
&\leq \sqrt{|| [\sum_{t
\geq t_>} \lambda_t \left\langle \phi_{t},y\right\rangle_{q} \vec{\phi}_t]||^2 ||[K+I \sigma^2]^{-1} \vec{y}||^2} 
\nonumber \\ 
&\leq \sigma^{-2} \sqrt{|| [\sum_{t\geq t_>} \lambda_t \left\langle \phi_{t},y\right\rangle_{q} \vec{\phi}_t]||^2 ||\vec{y}||^2} 
\nonumber \\
&= \lambda_> \sigma^{-2}\sqrt{ \sum_{\mu}\left[\sum_{t\geq t_>} \frac{\lambda_t}{\lambda_>} \left\langle \phi_{t},y\right\rangle_{q} \phi_t(x_{\mu})\right]^2 ||\vec{y}||^2}
\end{align}
Notably, if all $\lambda_t$'s are degenerate, we retrieve our previous bound with the feature being $\sum_t \left\langle \phi_{t},y\right\rangle_{q} \phi_t(x)$. More generally, we need to average the feature $\sum_t \frac{\lambda_t}{\lambda_>}\left\langle \phi_{t},y\right\rangle_{q} \phi_t(x)$ squared over the training set. 

Here we suggest several ways of treating this latter average. In some cases, where we have good control of all $\lambda_t$'s and $\phi_t$'s, we may know how to bound this last quantity directly. Alternatively, we may write 
\begin{align}
\sum_{\mu}&\left[\sum_{t\geq t_>} \frac{\lambda_t}{\lambda_>} \left\langle \phi_{t},y\right\rangle_{q} \phi_t(x_{\mu})\right]^2 
\nonumber \\
&= \sum_{t,p \geq t_>} \frac{\lambda_t}{\lambda_>} \frac{\lambda_p}{\lambda_>} \left\langle \phi_{t},y\right\rangle_{q} \left\langle \phi_{u},y\right\rangle_{q} \left[{\vec \phi}_t^{\,T} {\vec \phi}_{u}\right] 
\end{align}
and argue that in a typical scenario, ${\vec \phi}_t^{\, T} {\vec \phi}_{u}$ with $u \neq t$ would be smaller and also sum up incoherently. The diagonal contributions would thus be the dominant ones. Further assurance may be obtained by sampling $\lambda_t \lambda_u \left\langle \phi_{t},y\right\rangle_{q} \left\langle \phi_{u},y\right\rangle_{q} {\vec \phi}_t^{\,T} {\vec \phi}_u$ and verifying that off-diagonal contributions are indeed smaller and incoherent. Considering the diagonal contribution alone we obtain 
\begin{align}
|&\langle \hat{f}_D (x),y_>(x)\rangle|_q & 
\\
&\leq P\lambda_> (1-\epsilon) \sigma^{-2}\sqrt{\sum_{t\geq t_>} \left( \frac{\lambda_t}{\lambda_>} \right)^2 \left\langle \phi_{t},y\right\rangle_{q}^2 \E_{D} [\phi_t^2] \E_{D}[y^2]} 
\end{align}
which recalling $\lambda_t/\lambda_> \leq 1$ and $\sum_{t \geq t_>}[\left\langle \phi_{t},y\right\rangle_{q}]^2=O(1)$ yields a similar result to before.

Otherwise, we can take a worst-case scenario in which all $\phi_t(x_{\mu})$ contribute coherently to the sum [indeed think of $\phi_t$ as 1d Fourier modes, and the training distribution is a delta function at zero, and we wish to learn a delta function of the training set. Our $[K+\sigma^2]^{-1}y \approx \sigma^{-2} y$ estimate would be very poor however the $\phi_t$'s would all sum coherently around zero]. In this case, we may use Cauchy Schwarz again on the summation over $t$ to obtain 
\begin{align}
|&\langle \hat{f}_D (x),y_>(x)\rangle|_q &
\\
&\leq \lambda_> \sigma^{-2} (1-\epsilon) \sqrt{ \sum_{t\geq t_>} |\left\langle \phi_{t},y\right\rangle_{q}|^2 \sum_{\mu,t} \left|\frac{\lambda_t}{\lambda_>} \phi_t(x_{\mu})\right|^2 ||\vec{y}||^2} 
\end{align}
Consider this as a learnability namely $\langle f,y_>\rangle_q/\langle y_>,y_>\rangle_q$ bearing in mind that $y_>(x)$ has an $O(1)$ norm (or more) also under $q$. Scaling-wise, we may thus remove the $\sum_{t \geq t_>}|\left\langle \phi_{t},y\right\rangle_{q}|^2$ factor. Doing so we retrieve our previous bound, only with $\sum_t \frac{\lambda^2_t}{\lambda^2_>}E_D(\phi_t^2)$ instead of just $E_D(\phi_t^2)$. 

Notably, since in our normalization all $\lambda_t<1$, $\sum_t \lambda^2_t$ decays to zero quicker than $\lambda_t$ and hence by the finiteness of the trace yields a finite number even if $y_>$ contains an infinite amount of features. 

\section{Measures of Learnability \& Learning Parity with a Correlated Dataset}
\label{appendix:parity}
We begin by recovering a known result - learning parity from a uniform distribution on the hyper-cube with an FCN-GP is hard~\citep{yang_fine-grained_2020,simon_eigenlearning_2023} and extend it to general distributions on the hyper-cube. We then present an example where the data measure has very low entropy and probes only a small low dimensional space of the hyper-cube. In that example, a function that mostly agrees with parity on the data can be learned easily, but learning a predictor that generalizes out-of-distribution remains hard. We discuss what learning a function can mean when considering different distributions, potentially including out-of-distribution test points, and suggest maximal entropy distributions as a reasonable measure upon which learnability can be gauged. Finally, we compare the suggested measure of learnability to a familiar one.

Consider learning parity on the $\vec{x} \in \{-1,1\}^d$ hyper-cube using FCNs. Here $\vec{x}$ is drawn from an arbitrary, possibly correlated, measure on the hyper-cube, and the target function is parity $y=\Pi_{i=1}^d x_i$ with no added noise. 

Learning parity with noise is believed to require $P$ scaling exponentially with $d$ \citep{Blum2000}. Parity without noise can be learned with $O(d)$ samples using Gaussian elimination and relations between boolean operations and algebra in $Z_2$ fields \citep{Blum2000}. A GP, however, involves a larger hypothesis class including real rather than boolean variables. It also seems highly unlikely that it could learn from examples an $O(d^3)$-algorithm such as Gaussian elimination. In the case of a uniform measure, an FCN GP is known to require $P^*$ which is exponential in $d$ \citep{simon_eigenlearning_2023} to learn parity. It is reasonable to assume that a generic non-uniform measure would not reduce the complexity of this task, however, we are not aware of any existing bounds applicable to this broader case.

To this end, we take as an ideal distribution $q$ a uniform distribution on the sphere containing the corners of the hyper-cube.
We turn to calculate or bound the different elements in Eq. \ref{eq:main_result}. First, we require the $\phi_t(x)$ associated with parity.
Under $q$, any FCN kernel is diagonal in the basis of hyperspherical harmonics. The latter can be described as rank-$n$ homogeneous harmonic polynomials \citep{Frye2012}. Each rank constitutes an irreducible representation of the rotation group \citep{tung_group_1985,fulton_representation_2004}. As a consequence, each rank-$n$ polynomial is a kernel eigenfunction whose eigenvalue depends only on $n$.
We may conclude our eigenfuction of interest $\phi_t$ is $\hat{n}^{-1/2} y$ where $\hat{n}$ is a normalization factor w.r.t. to the measure $q$.
Having identified the eigenfunction we would like to estimate its eigenvalue.
One can show that there are $N(n,d)=\frac{2n+d-2}{n}\binom{n+d-3}{n-1}$ polynomials at given $n,d$ \citep{Frye2012}. Considering parity, it is a harmonic homogeneous polynomial of rank $d$ and consequently part of a $N(d,d)$-degenerate subspace of any FCN kernel with eigenvalue $\lambda_d$. Noting that $\E_{x\sim q} [k(x,x)]$ equal the sum of all eigenvalues and that eigenvalues are positive, we obtain $\lambda_d \leq \E_{x \sim q}[k(x,x)] /N(d,d)$. 

Following its appearance in the numerator and denominator and the fact that parity squares to 1 on the hyper-cube we find 
\begin{equation}
\begin{aligned}
  P^*
  &\geq
  \sigma^{2} \lambda_{t}^{-1} (1-\epsilon) \frac{\E_{x \sim q(x)} [y(x) y(x)]}{\sqrt{\E_{x\sim D}\left[y^{2}\left(x\right)\right]\E_{z\sim D}\left[y^{2} (x) \right]}} 
  \\
  &\geq \sigma^{2} (1-\epsilon) \hat{n} N(d,d)/\E_{x \sim q} [k(x,x)]
  \end{aligned}
\end{equation}

We calculate the normalization factor $\hat{n}$ in Appendix \ref{appendix:parity_fcn_norm}, and quote the result here
\begin{equation}
\hat{n} = 
\frac{2^{-d} d^{d} \Gamma \left(\frac{d}{2}\right)}{\Gamma \left(\frac{3 d}{2}\right)}.
\end{equation}
Next, we use Stirling's formula for an asymptotic expansion for $d \gg 1$
\begin{align}
P^* &\geq \frac{\sigma^2 (1-\epsilon)}{\E_{x \sim q}[k(x,x)]} \sqrt{\frac{3^{3}}{2^{6}\pi d}}\left(\frac{4e}{3^{3/2}}\right)^{d} 
\end{align}
We find that in high-dimension $d \gg 1$, given any training dataset on the hyper-cube, the sample complexity of parity for a FCN GP is at least exponential in $d$.

An extreme yet illustrative case to consider is a predominantly correlated measure on the hyper-cube which forces all $x_i$'s to be equal $p_1(x)$ plus a uniform measure namely $p(x) = (1-\alpha) p_1(x)+\alpha q(x)$ with $\alpha \ll 1$. For even $d$ on such measure, $y$ would be well approximated by a constant on $D$. While a constant function is easily learnable and may appear to yield a low test loss on $p(x)$, it grossly differs from the true target on $q$. Thus, in terms of generalization, the bound is useful for gauging the generalization properties on the analytically tractable measure $q$, rather than on the empirical measure ($p(x)$) on which the GP would seem to perform very well. In scenarios where a complex feature (such as parity) on the ideal measure $q$ is well approximated by much simpler ones (e.g. a constant) on the training measure one should be wary of associating an unlearnable target, in the sense of our bound, with poor generalization performance within the training distribution. 


\emph{Measuring learnability on $q$}. As the example above manifests, it is crucial to note our bound essentially bounds the learnability from the perspective of $q$ as can be seen in eq.\eqref{eq:our_learnability}. That means even for $P<P^*$ the model can perform well on the training dataset and even test datasets. Prominently, this can happen when training (resp. testing) on a low-entropy distribution that can collapse complicated functions onto simpler ones. 
In this sense, symmetric measures can be seen as maximal entropy distributions under certain constraints; they thus 
suggest themselves as a ground upon which OOD generalization can be predicted. For more complicated domains it remains an open question which dataset ``truly" reflects the feature $\phi_t(x)$ though we expect these differences to be small in practice. 

\section{Calculating the normalization factor for parity \& FCN}
\label{appendix:parity_fcn_norm}
To calculate this normalization factor we first extend it to be a function of the sphere radius namely
\begin{align}
\begin{aligned}
\hat{n}[r] &= 
\left(\frac{r^{d-1} 2 \pi^{d/2}}{\Gamma(d/2)}\right)^{-1} \int_{R^d} d^d x \delta(|x|-r) \prod_{i=1}^d x^2_i 
\\
&=:
\left(\frac{r^{d-1} 2 \pi^{d/2}}{\Gamma(d/2)}\right)^{-1} n[r],
\end{aligned}
\end{align}
where the first factor is the hypersphere surface area in $d$ dimensions under the assumption of even $d$. 

While we are interested in $n[r=\sqrt{d}]$ we instead first look at 
\begin{align}
N[s] &= \int_0^{\infty} dr e^{-s r^2}n[r] = \int_0^{\infty} d(r^2) e^{-s r^2}\frac{n[r]}{2r}
\end{align}
Notably $N[s]$ is then the Laplace transform of $n[r]/(2r)$ viewed as a function of $r^2$. Calculating it based on the second expression amounts to independent Gaussian integrations and yields 
\begin{align}
N[s] &= \int_{R^d} d^d x e^{- s|x|^2} \prod_i x^2_i = \left(\frac{\pi}{2^2 s^3}\right)^{d/2}
\end{align}
Inverting this Laplace transform we obtain 
\begin{align}
\frac{n[r]}{2r} &= \frac{2^{-d} \pi ^{d/2} r^{3 d-2}}{\Gamma \left(\frac{3 d}{2}\right)} \Rightarrow n[r] = \frac{2^{-(d-1)} \pi ^{d/2} r^{3 d-1}}{\Gamma \left(\frac{3 d}{2}\right)} \\ 
\nonumber 
\hat{n}[r] &= \left(\frac{r^{d-1} 2 \pi^{d/2}}{\Gamma(d/2)}\right)^{-1}n[r] = 
\frac{2^{-d} r^{2 d} \Gamma \left(\frac{d}{2}\right)}{\Gamma \left(\frac{3 d}{2}\right)}
\end{align}

\section{Eigenvalues for copying heads}
\label{appendix:copying_heads}
For the data distribution $q(X)$ we may use the results of~\citep{lavie_towards_2024} to characterize the eigenvalues and eigenvectors of the NNGP/NTK kernel of a transformer. The approach relies on the permutation symmetry between tokens in the same sample and uses representation theory to upper bound eigenvalues based on their degeneracy. 
Eigenvalues that belong to the same irreducible representation (irrep) $R$ and degenerate subspace $V_R^i$ can be bounded by the kernel's trace over the dimension of the irrep
\begin{equation}
  \lambda_{V_R^i} \leq \frac{\E_{x \sim q} [k(x,x)]}{{\rm dim}_{R}},
\end{equation}
where $\lambda_{V_R^i}$ is the ${\rm dim}_{R}$-fold degenerate eigenvalue of the subspace $V_R^i$, $E_{x \sim q} [k(x,x)]$ is the kernel's trace and ${\rm dim}_{R}$ is the dimension of the irrep. 

In the case of $q$ from the main text, we have a permutation symmetry in sequence space and an additional permutation symmetry in vocabulary space, since it is uniformly distributed. We can use this additional symmetry to identify the spaces $V_R^i$ mentioned above within the space of the irrep $R$ (i.e. within the space that includes the multiplicity of $R$). 

We thus set to decompose the target function into irreps of the symmetric group. As shown in \citep{lavie_towards_2024} linear functions (such as a copying head) are decomposed into two irreps of the symmetric group, ``trivial" and ``standard". We will look and the ``standard" component both in sequence space and in vocabulary space in order to capture the most unlearnable feature that is required for the copy head target.
\begin{equation}
\begin{aligned}
  \vec{\phi}_t^a(X) &= \frac{1}{z} \left( \vec{x}^{\,a-1} - \frac{1}{L}\sum_{b=1}^L \vec{x}^{\,b} - \frac{1}{V} \right) ; 
  \\
  z &= \sqrt{L} \sqrt{1-L^{-1}+L^{-1}V^{-1}} = \E_{X \sim q} [\vec{\phi}_t^{\,a}(X) \vec{y}^{\,a}(X)]
\end{aligned}
\end{equation}
It is part of a $(L-1)(V-1)$ degenerate space of the standard irrep of both the vocabulary and sequence permutation symmetry
\begin{equation}
  {\rm dim}_t = (L-1)(V-1)
\end{equation}
hence the eigenvalues can be bounded by
\begin{equation}
  \lambda_t \leq \frac{\E_{x \sim q} [k(x,x)]}{(L-1)(V-1)}.
\end{equation}


\end{document}